\documentclass[10pt,twocolumn,letterpaper]{article}

\usepackage{cvpr}              %

\input{CVPR2026/preamble}

\definecolor{cvprblue}{rgb}{0.21,0.49,0.74}
\usepackage[pagebackref,breaklinks,colorlinks,allcolors=cvprblue]{hyperref}

\def\paperID{43013} %
\def\confName{CVPR}
\def\confYear{2026}

\usepackage{xcolor}
\usepackage[normalem]{ulem} %
\usepackage{xspace}
\usepackage{bm}

\usepackage[accsupp]{axessibility}

\title{THOM: Generating Physically Plausible Hand-Object Meshes From Text}

\author{
Uyoung~Jeong${}^{1}$ \quad
Yihalem Yimolal Tiruneh${}^{1}$ \\
Hyung~Jin~Chang${}^{2}$ \quad
Seungryul~Baek${}^{1}$ \quad
Kwang~In~Kim${}^{3}$ \vspace{0.2cm}\\
${}^{1}$UNIST \quad ${}^{2}$University of Birmingham \quad  ${}^{3}$POSTECH
}

\begin{document}
\twocolumn[{
\maketitle
\begin{center}
\captionsetup{type=figure}
\includegraphics[width=0.98\linewidth]{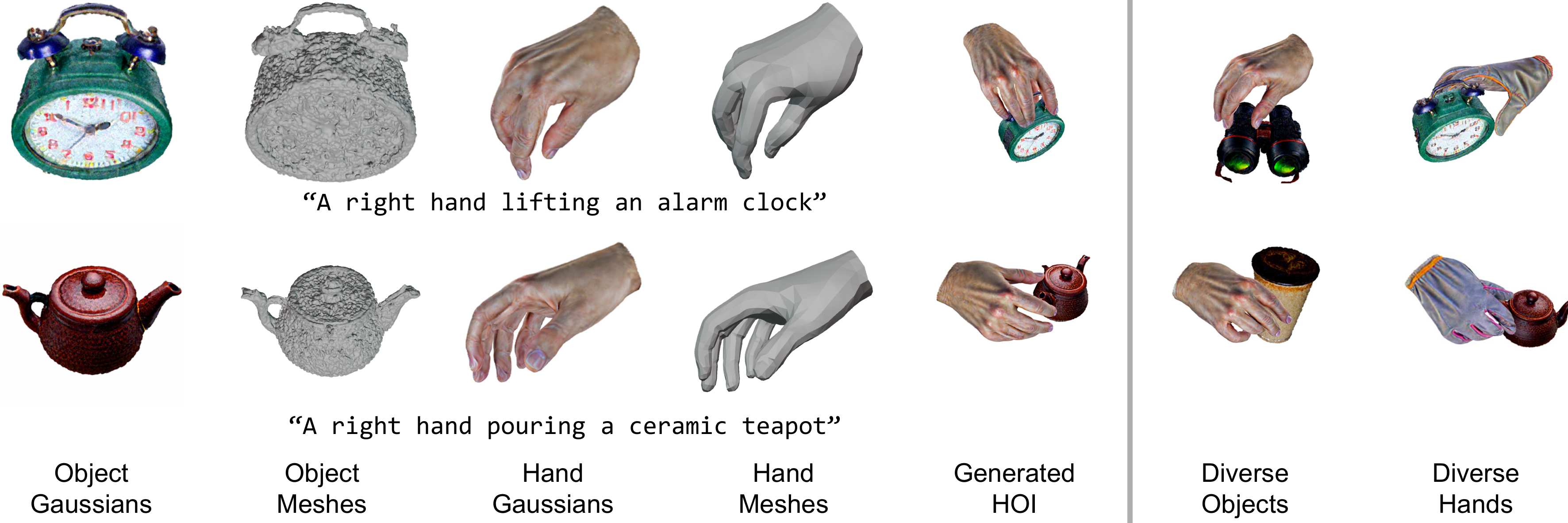}
\caption{\textbf{THOM} generates photorealistic and physically plausible 3D hand-object interactions from text prompts, with consistent hand and object meshes and realistic contact across diverse objects and hand appearances.
}
\label{fig:teaser}
\end{center}
}]

\begin{abstract}
Generating photorealistic 3D hand-object interactions (HOIs) from text is important for applications like robotic grasping and AR/VR content creation. In practice, however, achieving both visual fidelity and physical plausibility remains difficult, as mesh extraction from text-generated Gaussians is inherently ill-posed and the resulting meshes are often unreliable for physics-based optimization. We present THOM, a training-free framework that generates physically plausible 3D HOI meshes directly from text prompts, without requiring template object meshes. THOM follows a two-stage pipeline: it first generates hand and object Gaussians guided by text, and then refines their interaction using physics-based optimization. To enable reliable interaction modeling, we introduce a mesh extraction method with an explicit vertex-to-Gaussian mapping, which enables topology-aware regularization. We further improve physical plausibility through contact-aware optimization and vision-language model (VLM)-guided translation refinement. Extensive experiments show that THOM produces high-quality HOIs with strong text alignment, visual realism, and interaction plausibility.
\end{abstract}

\section{Introduction}
\label{sec:intro}
Generating photorealistic 3D hand-object interactions (HOIs) from text is a fundamental yet challenging task for dexterous robotic grasping~\cite{zhong2025dexgrasp, liang2025dexhanddiff, wang2025unigrasptransformer, li2025maniptrans}, collaborative robotics~\cite{zhao2025taste,liu2025core4d}, and AR/VR content creation~\cite{zhou2023mixed,kim2025shaping}.
It generates diverse object shapes and their corresponding hand-object interactions directly from text, without requiring reference 3D object assets. This significantly lowers the barrier for non-expert users to generate diverse interactions simply from text prompts.

Despite its application value, there are three key technical challenges that are unique to this text-based photorealistic HOI generation task:
(1) A lack of object and interaction diversity hinders the application of data-hungry learning-based approaches. While existing learning-based text-to-3D generation methods~\cite{tang2024lgm, wei2024meshlrm,xu2024instantmesh} are trained on the Objaverse dataset~\cite{deitke2023objaverse} containing $\sim$818K objects with 21K classes, 3D HOI datasets consist of a few dozen object classes~\cite{hampali2020honnotate,taheri2020grab,fan2023arctic}, which hinders training with large-scale generation models.
(2) Existing optimization-based HOI generation methods~\cite{cao2024avatargo,dai2024interfusion} use NeRF-based implicit volumes, lacking explicit mesh representations for physics-based HOI optimization.
(3) Existing methods~\cite{cha2024text2hoi,Muchen_LatentHOI} use physics-based HOI losses on watertight template object meshes, and are therefore incapable of optimizing non-watertight object meshes generated from text.

To address the aforementioned challenges, we propose \textbf{THOM}, a training-free framework that generates diverse photorealistic HOI meshes from text with high physical plausibility (\cref{fig:teaser}).
First, to overcome the training data shortage issue, we take a per-sample optimization strategy, leveraging the generalization power of a text-to-image diffusion model and the representational efficiency of 3D Gaussians. %
Second, for robust physics-based HOI refinement coupled with volumetric rendering, THOM establishes explicit vertex-to-Gaussian mapping that facilitates physics-based HOI optimization. To address the limitations of previous mesh extraction methods that involve complex iterative refinement processes, we design a simplified object mesh extraction process. For robust HOI optimization, we further extract a lightweight and concise object mesh for efficient and accurate vertex normal estimation.
Third, our robust physics-based HOI optimization process improves interaction plausibility on noisy text-generated object meshes. Our distance-adaptive masking strategy and reposition loss robustly align hand joints on the surfaces of diverse objects, thereby improving physical plausibility.

To further improve text-3D HOI alignment, we additionally refine hand translation using the high-level visual grounding capability of an existing vision-language model (VLM).
Instead of resource-intensive fine-tuning~\cite{cheng2024spatialrgpt,li2025bridgevla} or adding extra networks~\cite{li2025seeground,huang2025fireplace}, we utilize an existing open-source VLM without any training or modification. We prompt a VLM with rendered images of multiple candidates with different hand translations, and ask it to select the candidate with the best alignment.

Our contributions are summarized as follows:

\begin{itemize} 
\item We generate photorealistic 3D HOI meshes with diverse shapes directly from text prompts.
\item Our vertex-Gaussian mapping and simplified object mesh extraction enable accurate physics-based optimization.
\item We achieve high physical plausibility via VLM-guided translation refinement and physics-based optimization with distance-adaptive masking and joint reposition loss.
\item Comprehensive experiments demonstrate superior visual realism under diverse object and hand shapes, with improved physical plausibility compared to SOTA methods.
\end{itemize}

\section{Related work}
\paragraph{Text-to-3D object generation.}
Recent works on text-to-3D generation can be broadly categorized into three types: per-object optimization, multi-view 2D-to-3D lifting, and Large Reconstruction Model (LRM). DreamFusion~\cite{poole2022dreamfusion} and subsequent works~\cite{chen2023fantasia3d, metzer2023latent, lukoianov2024score, wang2023score} apply Score Distillation Sampling (SDS) to NeRF~\cite{mildenhall2021nerf} renderings using a 2D image diffusion model as a prior. More recent methods~\cite{chen2024text, yi2023gaussiandreamer, liang2024luciddreamer, tang2023dreamgaussian, yi2024gaussiandreamerpro} replace NeRF with Gaussian Splatting~\cite{kerbl20233d} and further improve the optimization speed and memory usage. However, they struggle to accurately reconstruct human anatomy, particularly hands, and are not tailored for physically plausible hand-object interactions.%

2D-to-3D lifting approaches divide the task into separate sub-tasks: 2D image generation and subsequent 3D lifting. Zero-1-to-3~\cite{liu2023zero}, CAT3D~\cite{gao2024cat3d} and related works~\cite{liu2023one, liu2024one, wen2025ouroboros3d} use multi-view images to obtain view-consistent 3D representations. SAM 3D~\cite{chen2025sam} and Hunyuan3D 2.5~\cite{lai2025hunyuan3d} further improve reconstruction fidelity with large-scale architectures and massive data. However, they fail to model articulated hands or produce physically plausible interactions.

LRM-based approaches~\cite{tang2024lgm, wei2024meshlrm, xu2024instantmesh, NEURIPS2024_6d09ef61,chen2025primx} adopt learning-based methods to generate 3D objects. MeshFormer~\cite{NEURIPS2024_6d09ef61} employs 3D sparse convolutions and transformers to generate 3D representations. However, these methods often require extensive training costs on large-scale data. Similar to the lifting methods, these approaches also struggle to generate complex articulated content and lack explicit modeling of physical contact and penetration.

\begin{figure*}[t]
\centerline{\includegraphics[width=1\textwidth]{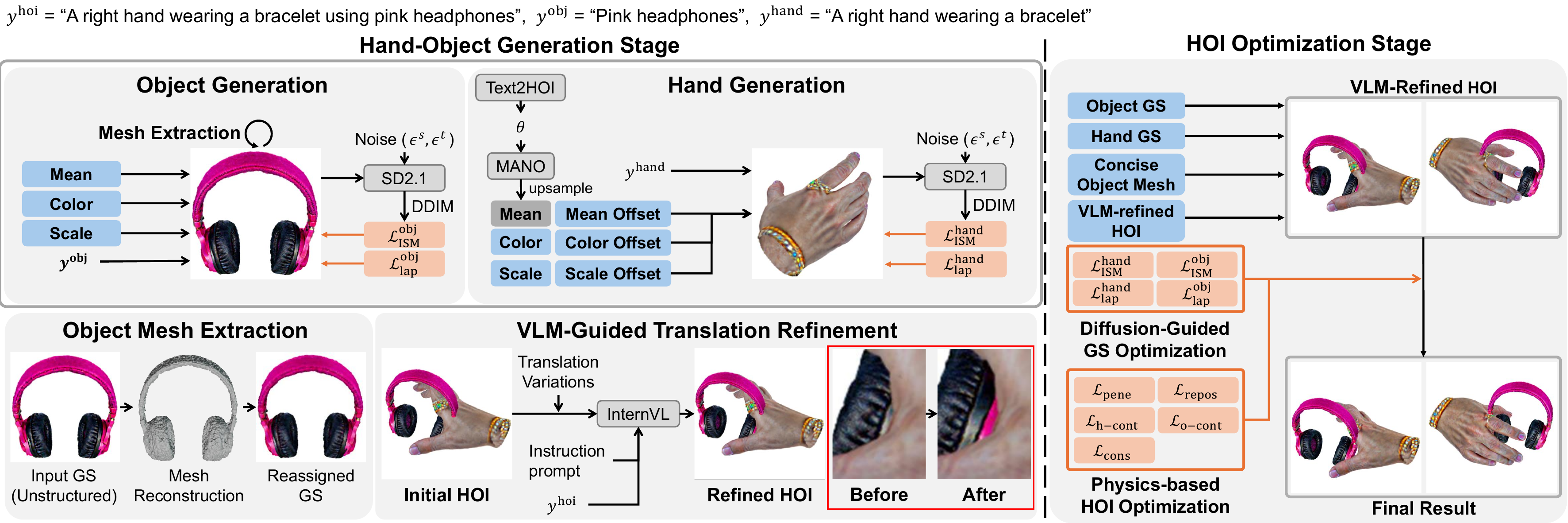}}
\caption{Overall pipeline of THOM. In the first stage, object and hand Gaussians are separately generated from the text prompts.
During the second stage, we jointly refine the HOI Gaussians and HOI parameters.
At HOI initialization, we refine the hand translation with VLM-guided refinement.
During HOI parameters optimization, we introduce physics-based optimization via distance-adaptive contact losses and reposition loss.
For both the first and the second stage, Laplacian regularization is applied for topological consistency.
}
\label{fig:thom_pipeline}
\end{figure*}

\paragraph{Human/hand-object interaction generation.}
Learning-based approaches~\cite{ye2023ghop, ghosh2023imos, cha2024text2hoi, zhang2025diffgrasp, diller2024cg} generate hand-object motions from the given object meshes. Although they offer a reasonable initialization prior, generalization to unseen object categories and shapes remains limited. In addition, they are unable to generate photorealistic scenes, and require template object meshes for HOI generation. On the other hand, our method generates photorealistic 3D Gaussians and their meshes from text without the need for the template object meshes and training data.

InterFusion~\cite{dai2024interfusion} and DreamHOI~\cite{zhu2024dreamhoi} take a per-instance optimization approach, generating 3D human-object interactions by optimizing NeRF volumes. However, these methods suffer from low-resolution renderings due to the computational cost of NeRF volumetric grid representation. Moreover, these approaches lack contact-awareness, resulting in physically implausible interactions.

There are several 2D HOI image generation approaches. Affordance Diffusion~\cite{ye2023affordance} and DiffusionHOI~\cite{zhang2024hoidiffusion} achieve this by training diffusion models on HOI datasets. While they are confined to generating 2D images and require computationally demanding training, our method generates photorealistic 3D HOIs without a learning-based approach.

\paragraph{Mesh extraction from volumetric representations.}
NeRF-based methods~\cite{wang2021neus, munkberg2022extracting, lin2023magic3d,  tang2023delicate} usually employ the iterative Marching Cubes algorithm to extract meshes, suffering from slow extraction speed and surface artifacts. Under Gaussian Splatting representation, SuGaR~\cite{guedon2024sugar} performs Poisson reconstruction~\cite{kazhdan2006poisson} by sampling point clouds from multi-view renderings, and binds Gaussians to triangles. Although faster than Marching Cubes, it requires heuristic tuning of the Gaussian counts, resulting in bumpy or incomplete surfaces and demanding extensive iterations. Our method, THOM, simplifies mesh extraction through direct Poisson reconstruction followed by vertex-based Gaussian upsampling, enabling explicit topological regularization on the Gaussians that preserves consistent 3D structure during optimization. For precise contact calculation, we additionally extract concise meshes from our reconstruction results.

\paragraph{VLMs for 3D reasoning.}
3D reasoning or grounding using VLMs is an actively studied field. Existing methods either finetune VLMs on task-specific data~\cite{cheng2024spatialrgpt,li2025bridgevla,sun2025layoutvlm,wang2025spatialclip} or introduce additional modules to refine VLM outputs~\cite{an2025generalized,zhang2024towards}, but both require annotated datasets and expensive training with limited generalizability. Recently, training-free methods have emerged~\cite{yuan2024visual,li2025seeground,huang2025fireplace} that directly utilize VLMs for 3D reasoning tasks. Motivated by these works, we propose a simple plug-and-play hand translation refinement strategy that requires no training or model modification.

\section{Method}
As illustrated in~\cref{fig:thom_pipeline}, our THOM framework adopts a two-stage pipeline for generating photorealistic 3D HOIs. Initially, object and hand meshes are independently generated with high visual realism. In the second stage, we jointly optimize their interaction parameters using physics-based losses for plausible contacts and minimal penetration.

\subsection{Diffusion guidance for 3D HOI generation}
To generate 3D HOIs using a pretrained text-to-2D image diffusion model, we employ Interval Score Matching (ISM) loss, a variant of Score Distillation Sampling (SDS) loss~\cite{liang2024luciddreamer}. ISM loss provides more realistic and detailed results compared to SDS loss.
Following GaussianDreamerPro~\cite{yi2024gaussiandreamerpro}, we compute the ISM gradient using a DDIM scheduler.
Given a diffusion model $\phi$, the gradient with respect to the Gaussian parameters $\psi$ is obtained as:
\begin{align}
\label{e:ismgradient}
\begin{split}
\nabla_{\psi} &\mathcal{L}_{\mathrm{ISM}}(\phi, \mathcal{R}) \triangleq \\
&\mathbb{E}_{t, \noise} \left[w(t) (\hat{\noise}_{\phi}(\mathcal{R}_{t}; y, t)-\hat{\noise}_{\phi}(\mathcal{R}_{s};\emptyset,s)) \frac{\partial \mathcal{R}}{\partial \psi}\right],
\end{split}
\end{align}
where $\mathcal{R}$ denotes the rendered Gaussian image, $w(t)$ is a time scheduling weight, $y$ is a text prompt, and $t\in[2, 980)$ is a timestep. $\hat{\noise}_{\phi}$ is the predicted noise (following LucidDreamer~\cite{liang2024luciddreamer}), $s(<t)$ is a timestep for inversion, and $\emptyset$ is an empty text prompt.

\subsection{Vertex-Gaussian model}\label{sec:mesh}
In hand-object generation, we adopt vertex-Gaussian mapping to maintain consistent topology that enables precise contact computation in the later HOI optimization stage.

\paragraph{Object Gaussian model.}
Given an object prompt $y^{\text{o}}$, the object Gaussians $G^{\text{o}}$ are optimized using the ISM gradient $\nabla_{\psi} \mathcal{L}_{\mathrm{ISM}}$ (\cref{e:ismgradient}). The Gaussians are parameterized by position $\gsmean\in\mathbb{R}^{N\times 3}$, scale $\mathbf{s}\in\mathbb{R}^{N\times 3}$, color $\mathbf{c}\in\mathbb{R}^{N\times 3}$, opacity $\mathbf{\alpha}\in\mathbb{R}^{N\times 1}$, and quaternion $\mathbf{q}\in\mathbb{R}^{N\times 4}$, where $N$ is the number of Gaussians. We control the number of Gaussians to balance rendering fidelity and computational efficiency. Prior to mesh reconstruction, our object Gaussians have at most $N=800{,}000$ Gaussians. After reconstruction, the Gaussians are reinitialized with the mesh vertices, resulting in $N^{\prime} \approx 400{,}000$ elements.

\paragraph{Hand Gaussian model.}\label{sec:hand}
Following the object Gaussian model, the hand Gaussian model $G^{\text{h}}$ is optimized using $\nabla_{\psi} \mathcal{L}_{\mathrm{ISM}}$ (\cref{e:ismgradient}) given a hand text prompt $y^{\text{h}}$.
Inspired by recent Gaussian avatar methods~\cite{hu2024expressive,pokhariya2024manus,moon2024exavatar}, we adopt the parametric MANO hand model~\cite{MANO2017} for initialization. 

Specifically, the MANO mesh is upsampled fourfold to 196,993 vertices. The Gaussian positions are computed via pose-dependent transformations defined as follows:
\begin{align}
\mathbf{M}^{\text{h}}(\manobeta, \manopose, \Delta\gsmean^{\text{h}}) &= W(\mathbf{T}(\manobeta, \manopose, \Delta\gsmean^{\text{h}}), \mathbf{J}(\manobeta), \manopose, \mathcal{W}),\label{eq:mano_lbs} \\
\mathbf{T}(\manobeta, \manopose, \Delta\gsmean^{\text{h}}) &= \overline{\mathbf{T}} + B_S(\manobeta) + B_P(\manopose) + \Delta\gsmean^{\text{h}},
\label{eq:mano_shape_transform}
\end{align}
where $W$ is a linear blend skinning function, $\mathbf{T}$ is a transformation of a template mesh $\overline{\mathbf{T}}$ by a given hand pose $\manopose$ and hand shape $\manobeta$ parameters. $B_S$ and $B_P$ are blend shape functions, $\mathbf{J}$ is a set of hand joints, and $\mathcal{W}$ is a blend weight matrix. We set $\manobeta$ to 0 for simplicity.
The position offset $\Delta\gsmean^{\text{h}}$ enables per-vertex deformation of the Gaussian positions from the original MANO mesh.

In addition to position offsets, we optimize for Gaussian color and scale, denoted as $\Delta\mathbf{c}$ and $\Delta\mathbf{s}$, respectively. The final color and scale of the Gaussians are computed by adding respective offsets: $\mathbf{c}^{\prime\text{h}} = \mathbf{c}^{\text{h}} + \Delta\mathbf{c}^{\text{h}}$, $\mathbf{s}^{\prime\text{h}} = \mathbf{s}^{\text{h}} + \Delta\mathbf{s}^{\text{h}}$.

\paragraph{Direct object mesh reconstruction.}
Existing mesh extraction methods struggle with text-generated object meshes and often require extensive refinement~\cite{wang2021neus, munkberg2022extracting,guedon2024sugar}. We simplify the process using Poisson reconstruction followed by vertex upsampling. Specifically, we upsample the object vertices to a target size of $N^{tgt}_{\text{obj}} = 400{,}000$, ensuring geometric alignment between mesh vertices and their corresponding Gaussian representations. To improve the robustness of vertex normal estimation on noisy text-generated meshes, we extract a concise mesh via farthest point sampling and alpha shape reconstruction~\cite{edelsbrunner2003shape}, which is used for the contact and reposition losses in~\cref{sec:hoi_opt}.

\paragraph{One-to-one vertex-Gaussian mapping.}
Our proposed mesh reconstruction preserves mesh topology, establishing a one-to-one vertex-Gaussian mapping by design.
While previous NeRF-based~\cite{lin2023magic3d,wang2023prolificdreamer} or triangle-bind~\cite{guedon2024sugar} approaches lack explicit geometric relationships between volumetric representations and the mesh, our method directly assigns each mesh vertex as a Gaussian.
This structural correspondence bridges 3DGS with mesh-based physics optimization, allowing accurate penetration and contact computation without any auxiliary processing.

\paragraph{Vertex-Gaussian topology regularization.}
Due to the under-constrained nature of SDS-based optimization, generated Gaussians can exhibit floating artifacts and inconsistent appearance. Our vertex-Gaussian mapping naturally enables topological regularization, which is widely used in meshes~\cite{liu2019soft,moon2024exavatar} but previously inapplicable to volumetric 3D representations.
Specifically, we apply Laplacian regularization on Gaussian positions $\gsmean$, colors $\mathbf{c}$, and scales $\mathbf{s}$, enforcing local smoothness based on the vertex topology:
\begin{equation}
\mathbf{L}(\mathbf{x}) =  \mathbf{x} - \sum_{k=1}^{K}\mathbf{x}_{(\cdot,j_{k})}\mathbf{w}_{(\cdot,k)},
\label{eq:lap}
\end{equation}
where $\mathbf{x} \in \{\gsmean, \mathbf{c}, \mathbf{s}\}$, $\mathbf{w}\in\mathbb{R}^{K}$ defines adjacency-weight vector for top-$K$ nearest neighbors, and $j_k$ denotes the index of the $k$-th nearest neighbor for each row.

For Gaussian color $\mathbf{c}$ and scale $\mathbf{s}$ parameters, we directly minimize their respective Laplacians to enforce local consistency. For Gaussian positions $\gsmean$, we match their Laplacian to that of the reference mesh vertices $\mathbf{V}$, preserving consistent topology. The overall Laplacian regularization loss is defined as:
\begin{equation}
\mathcal{L}_{\text{lap}} = \lambda_{\text{lap,}\gsmean} \left( \mathbf{L}(\gsmean) - \mathbf{L}(\mathbf{V}) \right)^{2} + \lambda_{\text{lap,c}} \mathbf{L}(\mathbf{c})^2 + \lambda_{\text{lap,s}} \mathbf{L}(\mathbf{s})^2,
\label{eq:lap_loss}
\end{equation}
where $\lambda_{\text{lap,}\gsmean}, \lambda_{\text{lap,c}}, \lambda_{\text{lap,s}}$ are the scalar weights.
This topological regularization is consistently employed throughout the optimization pipeline, excluding only the initial phase of object Gaussian generation prior to mesh reconstruction.

The optimization losses of the first stage for the object and the hand Gaussians are defined as follows:
\begin{align}
\mathcal{L}^{\text{o}}_{\text{stage1}} &=\mathcal{L}_{\text{ISM}}^{\text{o}}(\phi, \mathcal{R}^{\text{o}}) + \mathcal{L}_{\text{lap}}^{\text{o}}(\gsmean^{\text{o}}, \mathbf{V}^{\text{o}}, \mathbf{c}^{\text{o}}, \mathbf{s}^{\text{o}}), \\
\mathcal{L}^{\text{h}}_{\text{stage1}} &=\mathcal{L}_{\text{ISM}}^{\text{h}}(\phi, \mathcal{R}^{\text{h}}) + \mathcal{L}_{\text{lap}}^{\text{h}}(\gsmean^{\text{h}}, \mathbf{V}^{\text{h}}, \mathbf{c}^{\text{h}}, \mathbf{s}^{\text{h}}),
\end{align}
where $\mathcal{L}_{\text{lap}}^{\text{o}}$ is applied after the object mesh reconstruction.

\subsection{Hand-object interaction optimization}
\label{sec:hoi_opt}
\paragraph{HOI Gaussians initialization.}
Our final objective is to combine the generated hand and object Gaussians with plausible interaction. To construct a HOI Gaussian model $G^{\text{hoi}}$, we define HOI parameters $\hoiparam$ as below:
\begin{equation}
\hoiparam = \left(\mathbf{r}^{\text{hoi}}, \mathbf{t}^{\text{hoi}}, \manopose  \right),
\end{equation}
where $\mathbf{r}^{\text{hoi}}\in\mathbb{R}^{1\times 3}$ refers to the hand-relative object rotation in axis-angle format, $\mathbf{t}^{\text{hoi}}\in\mathbb{R}^{1\times 3}$ refers to the relative hand translation.
Subsequently, we define the positions of the HOI Gaussians as follows:
\begin{equation}
\gsmean^{\text{hoi}} = \left( s \gsmean^{\text{h}}(\mathbf{R}^{\text{hoi}})^{\top} + \mathbf{t}^{\text{hoi}}, \gsmean^{\text{o}} \right),
\end{equation}
where $\mathbf{R}^{\text{hoi}}$ is a $3\times 3$ rotation matrix obtained from $\mathbf{r}^{\text{hoi}}$. We similarly apply transformation on the object Gaussian scalings and rotations using HOI parameters. HOI parameters and the interacting hand pose are initialized with Text2HOI~\cite{cha2024text2hoi}. To optimize the Gaussians, we separately render the hand and object Gaussians without HOI parameters, and supervise with ISM loss. The total losses for HOI Gaussians optimization are defined as follows:
\begin{align}
\mathcal{L}_{\text{hoi-GS}} = \mathcal{L}_{\text{ISM}}^{\text{h}} + \mathcal{L}_{\text{ISM}}^{\text{o}} + \mathcal{L}_{\text{lap}}^{\text{h}}+\mathcal{L}_{\text{lap}}^{\text{o}},
\end{align}
where we omit the input arguments for simplicity. Note that we do not optimize ISM loss from the composited scene, since it leads to color mixing on the hand-object contact region, degrading fidelity.

\paragraph{VLM-guided HOI refinement.}
Although our physics-based HOI optimization in the later stage effectively improves physical plausibility, it does not ensure everything is in its right place, as it might produce misaligned hand positions by prioritizing penetration minimization without contextual understanding. To better align 3D and text with high-level HOI understanding, we utilize a vision-language model (VLM) for hand translation refinement before physics-based optimization.

Inspired by~\cite{huang2025fireplace}, we prompt an existing open-source VLM (InternVL3.5-14B~\cite{wang2025internvl3}) to select the most plausible hand translation from the set of rendered images.
Specifically, we generate 125($=5\times 5\times 5$) candidates by adding offsets in the range $\{-2, -1, 0, 1, 2\}$, each scaled by $1.0\times10^{-2}$, to the initial 3D translation along the three Cartesian axes. 
We then instruct the VLM to select the most plausible sample by comparing multiple candidates based on the rendered images.
For memory efficiency, we conduct the iterative selection process with a mini-batch of 3 images, repeating the process until only one candidate remains.

\begin{figure*}[t]
\begin{center}
\includegraphics[width=0.99\linewidth]{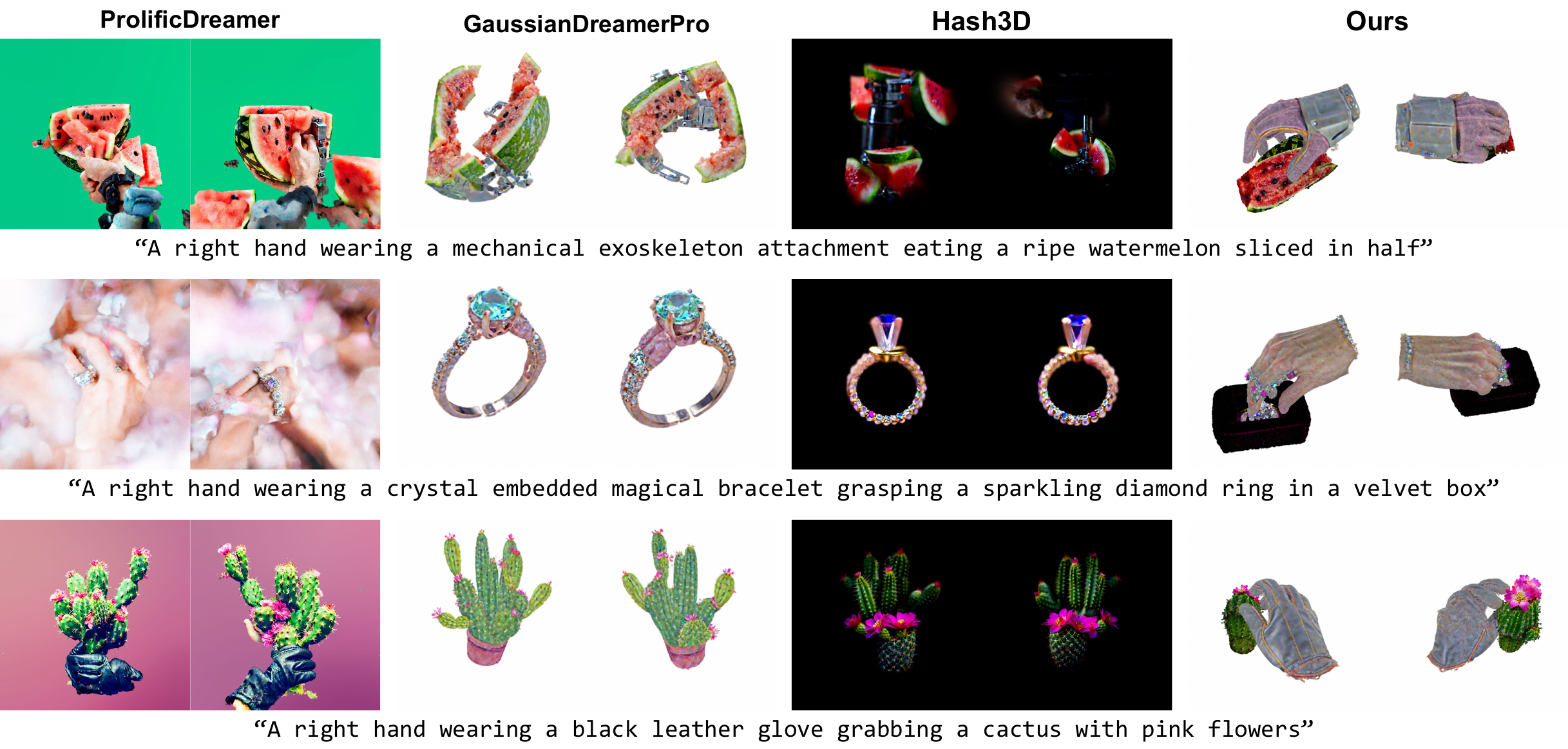}
\end{center}
\vspace{-2.5mm}
\caption{Qualitative comparisons of ProlificDreamer~\cite{wang2023prolificdreamer}, GaussianDreamerPro~\cite{yi2024gaussiandreamerpro}, Hash3D~\cite{yang2025hash3d} and ours.}
\label{fig:qualitative_comparison}
\end{figure*}

\paragraph{Physics-based HOI optimization.}
Thanks to our vertex-Gaussian representation and novel mesh extraction method, we can apply physics-based HOI optimization in our framework.
To optimize the HOI parameters, we use several regularization losses as defined below:
\begin{align} \label{eq:hoi_loss}
\mathcal{L}_{\text{hoi-phys}} &= \lambda_{\text{pene}}\mathcal{L}_{\text{pene}} + \lambda_{\text{hc}}\mathcal{L}_{\text{hc}} \\
&+\lambda_{\text{oc}} \mathcal{L}_{\text{oc}} + \lambda_{\text{repos}}\mathcal{L}_{\text{repos}} +  \lambda_{\text{cons}}\mathcal{L}_{\text{cons}},\nonumber
\end{align}
where these losses do not require rendering, as they are directly calculated from the geometric properties of the vertex-Gaussians. $\lambda$ represents the scalar weights for the respective losses. Extending from~\cite{jiang2021hand}, we use penetration loss and mask-based contact losses as defined below:
\begin{align}
\mathcal{L}_{\text{pene}}(\mathbf{n}^{\text{h}}, \gsmean^{\text{h}}, \gsmean^{\text{o}}) &= \|\gsmean^{\text{h}}_{\text{pene}} - \gsmean^{\text{o}}_{\text{pene}}\|_F^2, \\ \label{eq:pene_loss}
\mathcal{L}_{\text{oc}}(\mathbf{n}^{\text{h}}, \gsmean^{\text{h}}, \gsmean^{\text{o}}) &= \left\| \mathbf{C}_{\text{o}}\odot\left(\gsmean^{\text{o}}_{\text{oc}}-\gsmean^{\text{h}}_{\text{oc}} \right) \right\|_F^2, \\
\mathcal{L}_{\text{hc}}(\mathbf{n}^{\text{o}}_{\text{con.}}, \gsmean^{\text{o}}_{\text{con.}}, \gsmean^{\text{h}}) &= \left\| \mathbf{C}_{\text{h}}\odot\left(\gsmean^{\text{o}}_{\text{hc}}-\gsmean^{\text{h}}_{\text{hc}} \right) \right\|_F^2,
\label{eq:hand_cont}
\end{align}
where $\odot$ denotes the Hadamard product, $\|\cdot\|_F$ is the Frobenius norm. $\gsmean^{\text{o}}_{\text{pene}}\in\mathbb{R}^{N_{\text{pene}}\times 3}$ and $\gsmean^{\text{h}}_{\text{pene}}\in\mathbb{R}^{N_{\text{pene}}\times 3}$ represent the object Gaussian positions and their nearest hand counterparts in penetration, where $N_{\text{pene}}$ is the number of penetrating points. These are identified by computing the dot product between the hand vertex normals ($\mathbf{n}^{\text{h}}$) and $\mathbf{v}^{\text{o$\to$ h}} = \gsmean^{\text{h}} - \gsmean^{\text{o}}$.
$\gsmean^{\text{o}}_{\text{oc}}$ and $\gsmean^{\text{h}}_{\text{oc}}$ denote the object vertices and the nearest hand vertices. $\mathbf{C}_{\text{o}}$ is a binary mask for the object vertices where the contacting vertex indices (across $x,y,z$-dimensions) are set to 1.
$\gsmean^{\text{h}}_{\text{hc}}\in\mathbb{R}^{21\times 3}$ and $\gsmean^{\text{o}}_{\text{hc}}\in\mathbb{R}^{21\times 3}$ denote the hand joint positions and the nearest object vertices. Note that both $\mathbf{C}_{\text{o}}$ and $\mathbf{C}_{\text{h}}$ are initialized only once and do not change afterward.%

Object vertex contact loss $\mathcal{L}_{\text{oc}}$ minimizes the distance between the contacting object vertices and the nearest hand vertices.
During initialization of $\mathbf{C}_{\text{o}}$, we set the penetrating object vertices as contact points. If there is no penetration, the closest top-5 object vertices are marked as contact points. Compared to previous work~\cite{ye2023affordance,jian2023affordpose}, this loss effectively maps the contact affordance of the object without a time-consuming process.

Hand joint contact loss $\mathcal{L}_{\text{hc}}$ penalizes distance between the contact hand joints and the nearest object vertices.
During contact calculation, object vertex normals are required.
Existing works~\cite{jiang2021hand,cha2024text2hoi} assume watertight template meshes, thus failing on noisy text-generated meshes.
Therefore, we revisit $\mathcal{L}_{\text{hc}}$ in two aspects.
First, we use the concise object mesh for precise and robust object vertex normal estimation.
Second, we adapt the mask $\mathbf{C}_{\text{h}}$ to mark the top-5 hand keypoints closest to the object vertices as contact joints, improving joint-level contact alignment on noisy object surfaces.
These design choices effectively address the unique challenges of the text-to-HOI generation task.

\paragraph{Reposition loss.}
While the aforementioned losses improve overall physical plausibility, they do not explicitly relocate the penetrating hand joints to the object surface.  Instead, the optimization often pushes the hand away from the object without adjusting the hand pose.

To address this issue, we guide the penetrating hand joints to be closer to the nearest object surfaces when at least one of the following conditions is satisfied: (1) a joint is inside the object mesh; 
(2) a joint is marked as a contact joint.
For these joints, we compute L2 loss between the current MANO joint and the nearest object vertex positions, defined as follows:
\begin{equation}
\mathcal{L}_{\text{repos}}(\mathbf{n}^{\text{o}}_{\text{con.}}, \gsmean^{\text{o}}_{\text{con.}}, \mathbf{J}) = \left\| \mathbf{J}_{\text{repos}} - \gsmean^{\text{o}}_{\text{repos}} \right\|_F^2,
\end{equation}
where $\mathbf{J}_{\text{repos}},\gsmean^{\text{o}}_{\text{repos}}\in\mathbb{R}^{N_{\text{repos}}\times3}$, and $N_{\text{repos}}$ is the number of selected joints. Following~\cref{eq:hand_cont}, we use vertex normals of the object mesh to identify penetrating joints.

\paragraph{Consistency loss.}
$\mathcal{L}_{\text{cons}}$ regularizes the HOI parameters to prevent degenerate solutions:
\begin{align}
\begin{split}
\mathcal{L}_{\text{cons}}(\mathbf{t},\mathbf{r},\manopose) &= \|\mathbf{t}^{\text{hoi}} - \mathbf{t}^{\text{hoi}}_{\text{init}}\|^2 + \|\mathbf{r}^{\text{hoi}} - \mathbf{r}^{\text{hoi}}_{\text{init}}\|^2 \\
&+ \left\|\mathbf{W}_{xy} \odot(\manopose - \manopose_{\text{init}})\right\|_F^2,
\end{split}
\end{align}
where $\mathbf{t}^{\text{hoi}}_{\text{init}}$, $\mathbf{r}^{\text{hoi}}_{\text{init}}$, and $\manopose_{\text{init}}$ are the initial parameter values. $\mathbf{W}_{xy}$ applies a 10 times stronger penalization on the x- and y-axes, which govern left/right and horizontal joint rotations that are anatomically constrained, whereas the z-axis controls finger flexion and is less restricted.

\begin{table}[t]
  \centering
  \footnotesize
  \caption{Comparison with state-of-the-art text-to-3D and text-to-3D human-object interaction generation methods. *: Adapted to generate hand-object interactions.}
  \label{tab:benchmark_comparison}
  \resizebox{0.98\linewidth}{!}{
    \begin{tabular}{l|c|cc}
    \toprule
    Method & Time & CLIP $\uparrow$ & T$^{3}$-Align $\uparrow$\\
    \midrule
    DreamFusion~\cite{poole2022dreamfusion} & 35 min. & 27.1 & 1.3 \\
    ProlificDreamer~\cite{wang2023prolificdreamer}   & 3 h   & 29.1 & 1.2 \\
    GaussianDreamerPro~\cite{yi2024gaussiandreamerpro} & 1 h 30 min. & 28.5 & 1.4 \\
    Hash3D~\cite{yang2025hash3d} & 25 min. & 27.6 & 1.9 \\
    \midrule
    InterFusion*~\cite{dai2024interfusion} & 2 h & 22.5 & 1.5 \\
    DreamHOI*~\cite{zhu2024dreamhoi} & 3 h 20 min. & 27.8 & 1.6 \\
    \midrule
    Ours  & 2 h 30 min. & \textbf{31.4} & \textbf{2.6} \\
    \bottomrule
    \end{tabular}%
  }
\end{table}

\begin{figure*}[t]
\begin{center}
\includegraphics[width=1\linewidth]{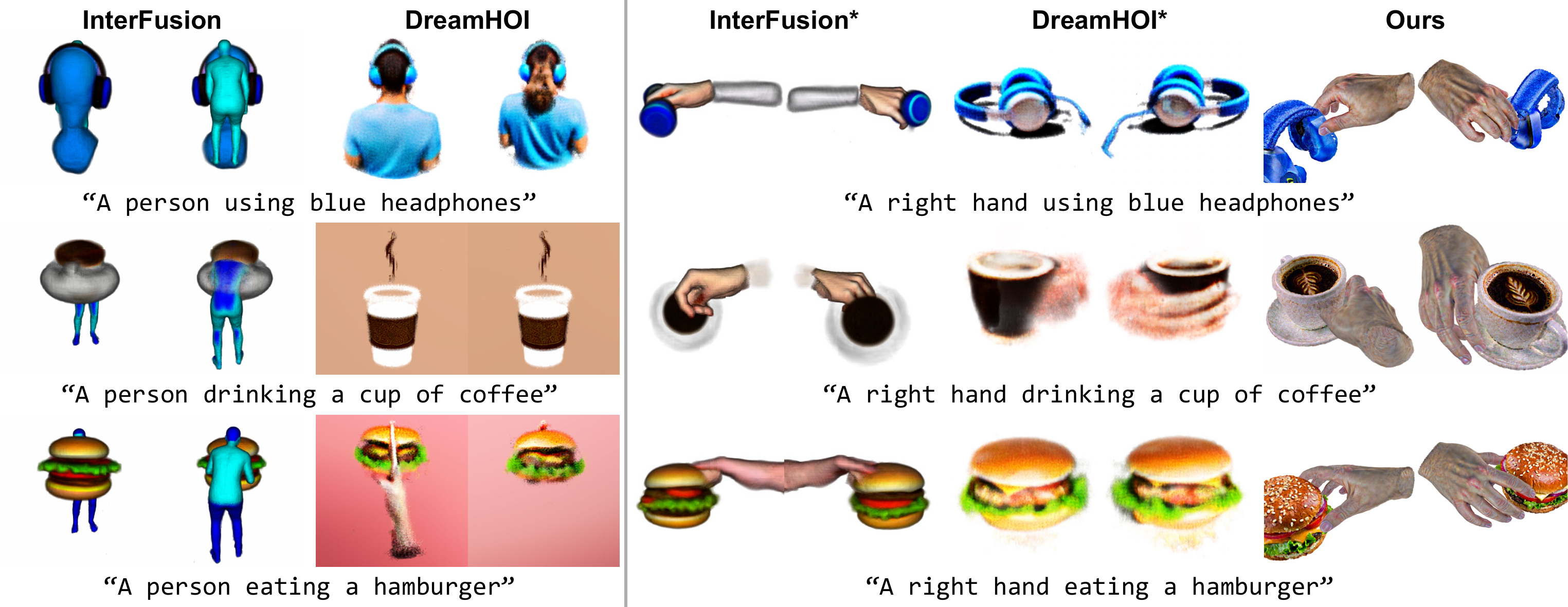}
\end{center}
\vspace{-2mm}
\caption{Qualitative comparison with human-object interaction methods. *: Adapted to generate hand-object interactions.
}
\label{fig:qualitative_comparison2}
\end{figure*}

\section{Experiments}
\subsection{Implementation details}
We generate 3D HOIs from input text prompts through two stages. \textbf{Stage 1}: Object and hand Gaussians are independently generated for 7,000 iterations. For the object Gaussians, mesh extraction is performed at the 5,000th iteration, followed by subsequent optimization with Laplacian regularization.
To guarantee visibility, we enforce a minimum opacity of 0.5 for object Gaussians and 1.0 for hand Gaussians.
\textbf{Stage 2}: HOI parameters are initialized using Text2HOI~\cite{cha2024text2hoi} and the translation parameter is refined with VLM. Then, the hand-object Gaussians and their HOI parameters are optimized for 1,000 iterations. The object Gaussian position parameters are frozen for stable contact refinement. The Gaussians and HOI parameters are separately optimized by different optimizers to avoid suboptimal visual quality due to physics-based objectives. Experiments are conducted on a single NVIDIA RTX 3090 GPU. The first stage takes approximately two hours, and the second stage takes around 30 minutes.

\begin{table}[!t]
\centering
\footnotesize
\centering
\setlength\tabcolsep{3pt}
\caption{Physics and simulation evaluation results with learning-based HOI generation methods.}
\resizebox{0.98\linewidth}{!}{
\begin{tabular}{l|cccc}
\toprule
Method & Max. pene. $\downarrow$ & Mean. pene. $\downarrow$ & Contact ratio $\uparrow$ & Disp. $\downarrow$ \\ \midrule
G-HOP & 1.30$\times 10^{-2}$ & 9.52$\times 10^{-3}$ & 0.15 & 1.00 \\
Text2HOI & 5.16$\times 10^{-4}$ & 2.28$\times 10^{-4}$ & \textbf{0.98} & 0.25 \\ \midrule
Ours & \textbf{2.20}$\times\bm{10^{-5}}$& \textbf{1.24}$\times\bm{10^{-5}}$ & 0.95 & \textbf{0.20} \\
\bottomrule
\end{tabular}
}
\label{table:main_phys_eval}
\end{table}

\subsection{Evaluation criteria}
To assess the generalization capability of our framework, we evaluate it on 100 different text prompts with 24 diverse action types. We use the object prompts from T$^{3}$Bench~\cite{he2023t3bench} for object generation, and we generate 100 hand prompts with diverse appearances using GPT-4o~\cite{hurst2024gpt}. Hand and object prompts are manually paired to ensure plausible interactions. Our evaluation set significantly exceeds previous benchmarks, such as InterFusion with 61 prompts and DreamHOI with 12 prompts.

We use CLIP score~\cite{radford2021learning} to evaluate the alignment between the input HOI text prompt and the rendered images, similar to the related works~\cite{dai2024interfusion,hu2024turbo3d}.
We further evaluate image-text alignment using T$^{3}$Bench Alignment metric~\cite{he2023t3bench}. While GPT-4V select metric used in InterFusion and DreamHOI is a relative evaluation protocol, T$^{3}$-Alignment metric is an absolute evaluation protocol and therefore independent of the comparative baselines.

\begin{table}[t]
  \centering
    \centering
    \footnotesize
    \caption{Ablation results with CLIP score. Higher scores indicate better performance. Our complete method shows the best score.}
    \label{tab:ablation}
    \resizebox{1\linewidth}{!}{
    \begin{tabular}{ccccc}
    \toprule
    w/o $\mathcal{L}_{\text{lap}}$ & w/o $\mathcal{L}_{\text{hoi-phys}}$ & w/o $\mathcal{L}_{\text{repos}}$ & w/o VLM & all \\ \midrule
    30.4 & 30.3 & 31.0 & 31.3 & \textbf{31.4} \\
    \bottomrule
    \end{tabular}
    }
\end{table}

\subsection{Results}
\paragraph{Quantitative results.}
To the best of our knowledge, there is no directly comparable method, as THOM is the first to generate \textit{photorealistic} 3D hand-object interactions directly from text prompts in a zero-shot setting. We therefore compare THOM with state-of-the-art text-to-3D and text-to-human-object interaction generation methods. \Cref{tab:benchmark_comparison} summarizes the results.

THOM outperforms prior approaches by a significant margin, indicating stronger alignment with interaction prompts. It also reduces generation time: while GaussianDreamerPro requires approximately three hours to generate two compositions, THOM achieves comparable results in about 30 minutes less while improving the CLIP score by 2.9 points. In comparison to ProlificDreamer, THOM improves both CLIP score (by 2.3 points) and T$^{3}$-Alignment (1.4 points). Furthermore, THOM consistently achieves the highest T$^{3}$-Alignment scores, outperforming all baselines by at least 0.7 points.

We compare our method with SOTA zero-shot human-object interaction generation methods. Since they are not directly applicable to hands, we adapt their implementations to generate hand-object interactions. Modification details are provided in the supplementary material. In~\cref{tab:benchmark_comparison}, we report quantitative comparison against InterFusion and DreamHOI. InterFusion and DreamHOI fail to generate high-fidelity and physically plausible results, as their implicit representations hinder accurate physics optimization. Our method significantly outperforms on both CLIP score and T$^{3}$-Alignment score, indicating that our method excels at generating 3D HOIs.

To validate physical plausibility of our generation results, we compare with learning-based HOI mesh generation methods in~\cref{table:main_phys_eval}. G-HOP and Text2HOI are trained from template object meshes and generate non-photorealistic HOIs, showing suboptimal performance on unseen text-generated object meshes. THOM significantly improves penetration by orders of magnitude, achieving the smallest penetration depths while maintaining a high contact ratio. Our method also effectively improves the displacement measured by PyBullet physics simulation.

\begin{table}[t]
    \centering
    \setlength\tabcolsep{4pt}
    \captionof{table}{Ablation of object mesh reconstruction methods.} 
    \label{tab:ablation_mesh_recon}
    \begin{tabular}{ccc}
        \toprule
        Method & CLIP $\uparrow$ & Pene. $\downarrow$ \\ \midrule
        Downsample & 30.9 & 7.3$\times10^{-5}$ \\
        Marching Cubes & 26.8 & 2.3$\times 10^{-5}$ \\
        SuGaR & 30.4 & 3.8$\times 10^{-5}$ \\
        Ours & \textbf{31.4} & \textbf{2.2}$\times \bm{10^{-5}}$ \\
        \bottomrule
    \end{tabular}
\end{table}

\paragraph{Qualitative results.}
Our method produces superior qualitative results.
In~\cref{fig:qualitative_comparison}, all state-of-the-art text-to-3D methods fail to generate hands with correct geometry.
Specifically, ProlificDreamer suffers from background clutter and a severe multi-face problem.
GaussianDreamerPro and Hash3D often fail to generate objects and hands separately, producing implausible results.
In contrast, our method consistently generates high-fidelity 3D HOIs, with plausible interactions, showing a strong generalization capability.

In \cref{fig:qualitative_comparison2}, we qualitatively compare with InterFusion and DreamHOI. Both InterFusion and DreamHOI fail to generate physically plausible interactions, due to the lack of explicit mesh representation and contact-awareness. InterFusion suffers from low-fidelity and cluttering artifacts. DreamHOI often fails to correctly generate hands, due to clutter in implicit NeRF volumes. On the other hand, our method consistently provides high-fidelity visual results along with plausible interactions.

\subsection{Ablation study}
In~\cref{tab:ablation}, we demonstrate the effectiveness of our proposed methods.
Removing the Laplacian loss (w/o $\mathcal{L}_{\text{lap}}$) significantly drops the CLIP score by 1.0. It degrades the topological consistency of object and hand meshes, resulting in degraded colors and jittery surfaces. Omitting all physics-based losses (w/o $\mathcal{L}_{\text{hoi-phys}}$) corresponds to using Text2HOI output without modification, resulting in a 1.1 point drop. The initial Text2HOI results suffer from significant interpenetration, undermining interaction plausibility. Removing only the reposition loss (w/o $\mathcal{L}_{\text{repos}}$) also leads to a suboptimal result, implying that reposition loss helps supervise the penetrating hand joints explicitly. We observe that the removal of VLM-guided refinement (w/o VLM) also drops the CLIP score by 0.1. VLM refinement largely improves T$^{3}$-Align by 0.6 points compared to the non-VLM baseline (details are provided in the supplementary material).%

\begin{table}[t]
    \centering%
    \caption{Ablation of physics optimization and VLM refinement using GPT-4o select metric. Values indicate the preference frequency (one sample excluded due to GPT-4o non-response).}
    \label{tab:ablation_gpt4o_select}
    \setlength{\tabcolsep}{4pt}
    \begin{tabular}{ccc}
    \toprule
    w/o phys. opt & w/o VLM & Ours (all) \\ \midrule
    13 & 25 & \textbf{61} \\
    \bottomrule
    \end{tabular}
\end{table}

\begin{figure}[t]
\begin{center}
\setlength{\tabcolsep}{.1pt}
\resizebox{0.98\linewidth}{!}{
\begin{tabular}{lccc}
\shortstack{Before\\refine} & \raisebox{-0.5\height}{\includegraphics[height=0.245\linewidth]{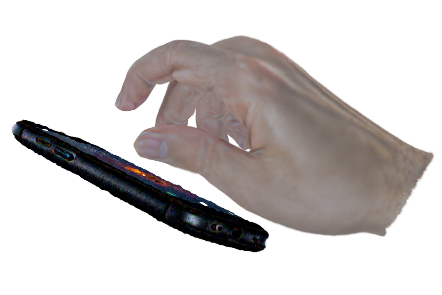}} & \raisebox{-0.5\height}{\includegraphics[height=0.245\linewidth]{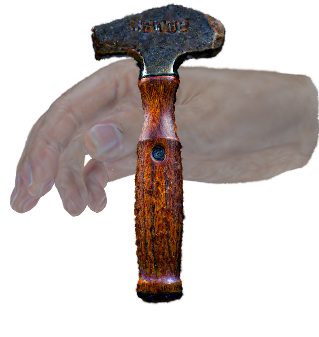}} & \raisebox{-0.5\height}{\includegraphics[height=0.245\linewidth]{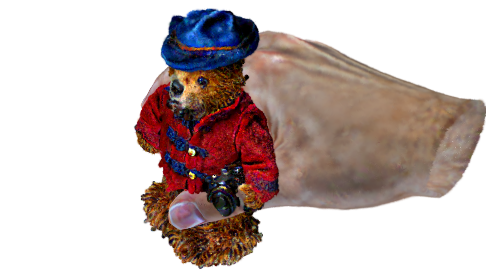}} \\
\shortstack{After\\refine} & \raisebox{-0.5\height}{\includegraphics[height=0.245\linewidth]{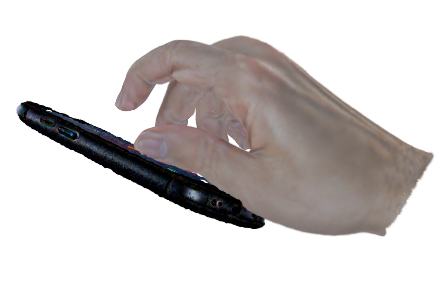}} & \raisebox{-0.5\height}{\includegraphics[height=0.245\linewidth]{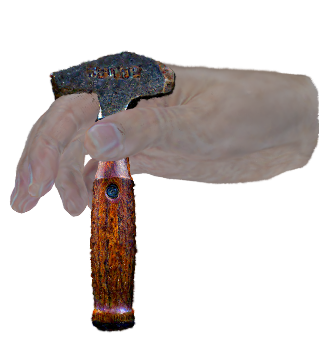}} & \raisebox{-0.5\height}{\includegraphics[height=0.245\linewidth]{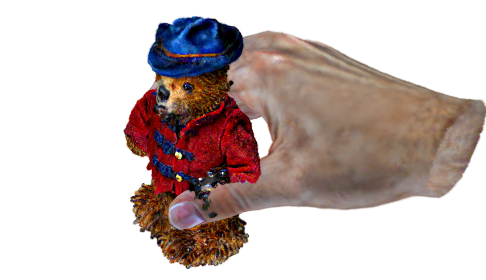}} \\
\end{tabular}
}
\end{center}
\vspace{-2mm}
\caption{VLM refinement results. Left: ``A right hand calling a smartphone''. Middle: ``A right hand using a hammer''. Right: ``A right hand inspecting Paddington Bear''.}
\label{fig:vlm_selection_example}
\end{figure}

We also provide ablation results on our mesh extraction method in~\cref{tab:ablation_mesh_recon}. We replace our mesh extraction method with other methods and then apply HOI optimization. For Marching Cubes, we use point-E~\cite{nichol2022point}. Our proposed mesh extraction method surpasses other baselines in CLIP score and maximum penetration. Downsampling, which downsamples from the baseline SuGaR, shows the worst penetration, as it exhibits severe non-watertight topology. While artifacts in Marching Cubes and jittery triangle-bound SuGaR hinder accurate penetration and contact estimation, our method empowers robust HOI optimization.

In~\cref{tab:ablation_gpt4o_select}, we isolate the effects of physics optimization and VLM refinement. From 100 generation results, we ask GPT-4o to select the preferred generation result among 3 candidates. Directly using the Text2HOI results (w/o phys. opt) was the least preferred. Incorporating both the physics optimization and VLM refinement is the most preferred method, validating the effectiveness of our framework.

\Cref{fig:vlm_selection_example} shows qualitative results of VLM-guided refinement. The refinement moves the hand toward plausible contact (left), improves grasp alignment on the handle (middle), and reduces thumb--object interpenetration (right), demonstrating consistent improvements in interaction plausibility across diverse objects.

\section{Conclusion}
We presented THOM, a training-free framework for generating photorealistic and physically plausible 3D hand-object interactions directly from text prompts. By combining a two-stage optimization pipeline with explicit vertex–Gaussian mapping, our method enables reliable mesh extraction and topology-aware regularization, allowing effective integration of physics-based interaction optimization within a volumetric representation.

Our approach further incorporates contact-aware losses, joint repositioning, and VLM-guided translation refinement, which together improve interaction plausibility and alignment with text prompts. Extensive experiments demonstrate that THOM achieves strong performance in terms of visual realism, text alignment, and physical consistency across diverse objects and hand configurations.

\paragraph{Limitations.} Despite these improvements, THOM still relies on per-instance optimization and therefore requires longer generation time than learning-based approaches. Future work could explore more efficient diffusion guidance or hybrid training-based strategies.

\section*{Acknowledgments}
This work was supported by the Institute of Information \& Communications Technology Planning \& Evaluation (IITP) grants (No. RS-2020-II201336, Artificial Intelligence Graduate School Program (UNIST); No. RS-2019-II191906, Artificial Intelligence Graduate School Program (POSTECH); No. RS-2022-II220290, Visual Intelligence for Space-Time Understanding and Generation; No. RS-2022-II220608, Artificial intelligence research about multi-modal interactions for empathetic conversations with humans), the National Research Foundation of Korea (NRF) grant (No. RS-2024-00337559), and the InnoCORE program (26-InnoCORE-01) of the Ministry of Science and ICT (MSIT), all funded by the Korea government (MSIT).
{
    \small
    \bibliographystyle{ieeenat_fullname}
    \bibliography{BibData}

@String(CVPR= {IEEE Conf. Comput. Vis. Pattern Recog.})

@String(ICCV= {Int. Conf. Comput. Vis.})

@String(ECCV= {Eur. Conf. Comput. Vis.})

@String(NIPS= {Adv. Neural Inform. Process. Syst.})

@String(AAAI = {AAAI})

@String(VR   = {Vis. Res.})

@String(CVPR = {CVPR})

@String(ICCV  = {ICCV})

@String(ECCV  = {ECCV})

@String(NIPS  = {NeurIPS})

@String(ICML = {ICML})

@String(aaai = {AAAI})

@article{poole2022dreamfusion,
  title={Dreamfusion: Text-to-3d using 2d diffusion},
  author={Poole, Ben and Jain, Ajay and Barron, Jonathan T and Mildenhall, Ben},
  journal={arXiv preprint arXiv:2209.14988},
  year={2022}
}

@inproceedings{lin2023magic3d,
  title={Magic3d: High-resolution text-to-3d content creation},
  author={Lin, Chen-Hsuan and Gao, Jun and Tang, Luming and Takikawa, Towaki and Zeng, Xiaohui and Huang, Xun and Kreis, Karsten and Fidler, Sanja and Liu, Ming-Yu and Lin, Tsung-Yi},
  booktitle=CVPR,
  pages={300--309},
  year={2023}
}

@article{lukoianov2024score,
  title={Score distillation via reparametrized ddim},
  author={Lukoianov, Artem and S{\'a}ez de Oc{\'a}riz Borde, Haitz and Greenewald, Kristjan and Guizilini, Vitor and Bagautdinov, Timur and Sitzmann, Vincent and Solomon, Justin M},
  journal=NIPS,
  volume={37},
  pages={26011--26044},
  year={2024}
}

@inproceedings{wang2023prolificdreamer,
  title={Prolific{D}reamer: High-Fidelity and Diverse Text-to-3D Generation with Variational Score Distillation},
  author={Zhengyi Wang and Cheng Lu and Yikai Wang and Fan Bao and Chongxuan Li and Hang Su and Jun Zhu},
  booktitle=NIPS,
  year={2023}
}

@inproceedings{yi2023gaussiandreamer,
  title={Gaussian{D}reamer: Fast Generation from Text to 3D Gaussians by Bridging 2D and 3D Diffusion Models},
  author={Yi, Taoran and Fang, Jiemin and Wang, Junjie and Wu, Guanjun and Xie, Lingxi and Zhang, Xiaopeng and Liu, Wenyu and Tian, Qi and Wang, Xinggang},
  year = {2024},
  booktitle = CVPR
}

@inproceedings{chen2024text,
  title={Text-to-3d using gaussian splatting},
  author={Chen, Zilong and Wang, Feng and Wang, Yikai and Liu, Huaping},
  booktitle=CVPR,
  pages={21401--21412},
  year={2024}
}

@article{tang2023dreamgaussian,
  title={Dreamgaussian: Generative gaussian splatting for efficient 3d content creation},
  author={Tang, Jiaxiang and Ren, Jiawei and Zhou, Hang and Liu, Ziwei and Zeng, Gang},
  journal={arXiv preprint arXiv:2309.16653},
  year={2023}
}

@article{yi2024gaussiandreamerpro,
  title={Gaussiandreamerpro: Text to manipulable 3d gaussians with highly enhanced quality},
  author={Yi, Taoran and Fang, Jiemin and Zhou, Zanwei and Wang, Junjie and Wu, Guanjun and Xie, Lingxi and Zhang, Xiaopeng and Liu, Wenyu and Wang, Xinggang and Tian, Qi},
  journal={arXiv preprint arXiv:2406.18462},
  year={2024}
}

@inproceedings{guedon2024sugar,
  title={Sugar: Surface-aligned gaussian splatting for efficient 3d mesh reconstruction and high-quality mesh rendering},
  author={Gu{\'e}don, Antoine and Lepetit, Vincent},
  booktitle=CVPR,
  pages={5354--5363},
  year={2024}
}

@inproceedings{chen2023fantasia3d,
  title={Fantasia3d: Disentangling geometry and appearance for high-quality text-to-3d content creation},
  author={Chen, Rui and Chen, Yongwei and Jiao, Ningxin and Jia, Kui},
  booktitle=ICCV,
  pages={22246--22256},
  year={2023}
}

@inproceedings{metzer2023latent,
  title={Latent-nerf for shape-guided generation of 3d shapes and textures},
  author={Metzer, Gal and Richardson, Elad and Patashnik, Or and Giryes, Raja and Cohen-Or, Daniel},
  booktitle=CVPR,
  pages={12663--12673},
  year={2023}
}

@inproceedings{wang2023score,
  title={Score jacobian chaining: Lifting pretrained 2d diffusion models for 3d generation},
  author={Wang, Haochen and Du, Xiaodan and Li, Jiahao and Yeh, Raymond A and Shakhnarovich, Greg},
  booktitle=CVPR,
  pages={12619--12629},
  year={2023}
}

@article{mildenhall2021nerf,
  title={Nerf: Representing scenes as neural radiance fields for view synthesis},
  author={Mildenhall, Ben and Srinivasan, Pratul P and Tancik, Matthew and Barron, Jonathan T and Ramamoorthi, Ravi and Ng, Ren},
  journal={Communications of the ACM},
  volume={65},
  number={1},
  pages={99--106},
  year={2021},
  publisher={ACM New York, NY, USA}
}

@article{kerbl20233d,
  title={3d gaussian splatting for real-time radiance field rendering.},
  author={Kerbl, Bernhard and Kopanas, Georgios and Leimk{\"u}hler, Thomas and Drettakis, George},
  journal={ACM Trans. Graph.},
  volume={42},
  number={4},
  pages={139--1},
  year={2023}
}

@inproceedings{ghosh2023imos,
  title={{IM}o{S}: Intent-Driven Full-Body Motion Synthesis for Human-Object Interactions},
  author={Ghosh, Anindita and Dabral, Rishabh and Golyanik, Vladislav and Theobalt, Christian and Slusallek, Philipp},
  booktitle={Computer Graphics Forum},
  volume={42},
  pages={1--12},
  year={2023},
  organization={Wiley Online Library}
}

@inproceedings{cha2024text2hoi,
  title={Text2{HOI}: Text-guided 3d motion generation for hand-object interaction},
  author={Cha, Junuk and Kim, Jihyeon and Yoon, Jae Shin and Baek, Seungryul},
  booktitle=CVPR,
  pages={1577--1585},
  year={2024}
}

@inproceedings{zhang2025diffgrasp,
  title={Diffgrasp: Whole-body grasping synthesis guided by object motion using a diffusion model},
  author={Zhang, Yonghao and He, Qiang and Wan, Yanguang and Zhang, Yinda and Deng, Xiaoming and Ma, Cuixia and Wang, Hongan},
  booktitle=AAAI,
  volume={39},
  pages={10320--10328},
  year={2025}
}

@inproceedings{diller2024cg,
  title={Cg-hoi: Contact-guided 3d human-object interaction generation},
  author={Diller, Christian and Dai, Angela},
  booktitle=CVPR,
  pages={19888--19901},
  year={2024}
}

@inproceedings{dai2024interfusion,
  title={Inter{F}usion: Text-Driven Generation of 3D Human-Object Interaction},
  author={Dai, Sisi and Li, Wenhao and Sun, Haowen and Huang, Haibin and Ma, Chongyang and Huang, Hui and Xu, Kai and Hu, Ruizhen},
  booktitle=ECCV,
  year={2024}
}

@inproceedings{radford2021learning,
  title={Learning transferable visual models from natural language supervision},
  author={Radford, Alec and Kim, Jong Wook and Hallacy, Chris and Ramesh, Aditya and Goh, Gabriel and Agarwal, Sandhini and Sastry, Girish and Askell, Amanda and Mishkin, Pamela and Clark, Jack and others},
  booktitle=ICML,
  pages={8748--8763},
  year={2021},
  organization={PmLR}
}

@inproceedings{liu2023zero,
  title={Zero-1-to-3: Zero-shot one image to 3d object},
  author={Liu, Ruoshi and Wu, Rundi and Van Hoorick, Basile and Tokmakov, Pavel and Zakharov, Sergey and Vondrick, Carl},
  booktitle=ICCV,
  pages={9298--9309},
  year={2023}
}

@article{liu2023one,
  title={One-2-3-45: Any single image to 3d mesh in 45 seconds without per-shape optimization},
  author={Liu, Minghua and Xu, Chao and Jin, Haian and Chen, Linghao and Varma T, Mukund and Xu, Zexiang and Su, Hao},
  journal=NIPS,
  volume={36},
  pages={22226--22246},
  year={2023}
}

@inproceedings{liu2024one,
  title={One-2-3-45++: Fast single image to 3d objects with consistent multi-view generation and 3d diffusion},
  author={Liu, Minghua and Shi, Ruoxi and Chen, Linghao and Zhang, Zhuoyang and Xu, Chao and Wei, Xinyue and Chen, Hansheng and Zeng, Chong and Gu, Jiayuan and Su, Hao},
  booktitle=CVPR,
  pages={10072--10083},
  year={2024}
}

@inproceedings{liang2024luciddreamer,
  title={Luciddreamer: Towards high-fidelity text-to-3d generation via interval score matching},
  author={Liang, Yixun and Yang, Xin and Lin, Jiantao and Li, Haodong and Xu, Xiaogang and Chen, Yingcong},
  booktitle=CVPR,
  pages={6517--6526},
  year={2024}
}

@article{gao2024cat3d,
    title={{CAT3D}: Create Anything in 3D with Multi-View Diffusion Models},
    author={Ruiqi Gao* and Aleksander Holynski* and Philipp Henzler and Arthur Brussee and Ricardo Martin-Brualla and Pratul P. Srinivasan and Jonathan T. Barron and Ben Poole*
    },
    journal=NIPS,
    year={2024}
}

@inproceedings{tang2024lgm,
  title={Lgm: Large multi-view gaussian model for high-resolution 3d content creation},
  author={Tang, Jiaxiang and Chen, Zhaoxi and Chen, Xiaokang and Wang, Tengfei and Zeng, Gang and Liu, Ziwei},
  booktitle=ECCV,
  pages={1--18},
  year={2024},
  organization={Springer}
}

@article{wei2024meshlrm,
  title={Mesh{LRM}: Large Reconstruction Model for High-Quality Meshes},
  author={Wei, Xinyue and Zhang, Kai and Bi, Sai and Tan, Hao and Luan, Fujun and Deschaintre, Valentin and Sunkavalli, Kalyan and Su, Hao and Xu, Zexiang},
  journal={arXiv preprint arXiv:2404.12385},
  year={2024}
}

@article{xu2024instantmesh,
  title={Instantmesh: Efficient 3d mesh generation from a single image with sparse-view large reconstruction models},
  author={Xu, Jiale and Cheng, Weihao and Gao, Yiming and Wang, Xintao and Gao, Shenghua and Shan, Ying},
  journal={arXiv preprint arXiv:2404.07191},
  year={2024}
}

@article{NEURIPS2024_6d09ef61,
  title={Mesh{F}ormer : High-Quality Mesh Generation with 3D-Guided Reconstruction Model},
  author={Liu, Minghua and Zeng, Chong and Wei, Xinyue and Shi, Ruoxi and Chen, Linghao and Xu, Chao and Zhang, Mengqi and Wang, Zhaoning and Zhang, Xiaoshuai and Liu, Isabella and Wu, Hongzhi and Su, Hao},
  journal=NIPS,
  volume={37},
  pages={59314--59341},
  year={2024}
}

@inproceedings{tang2023delicate,
  title={Delicate textured mesh recovery from nerf via adaptive surface refinement},
  author={Tang, Jiaxiang and Zhou, Hang and Chen, Xiaokang and Hu, Tianshu and Ding, Errui and Wang, Jingdong and Zeng, Gang},
  booktitle=ICCV,
  pages={17739--17749},
  year={2023}
}

@inproceedings{munkberg2022extracting,
  title={Extracting triangular 3d models, materials, and lighting from images},
  author={Munkberg, Jacob and Hasselgren, Jon and Shen, Tianchang and Gao, Jun and Chen, Wenzheng and Evans, Alex and M{\"u}ller, Thomas and Fidler, Sanja},
  booktitle=CVPR,
  pages={8280--8290},
  year={2022}
}

@article{wang2021neus,
  title={Neus: Learning neural implicit surfaces by volume rendering for multi-view reconstruction},
  author={Wang, Peng and Liu, Lingjie and Liu, Yuan and Theobalt, Christian and Komura, Taku and Wang, Wenping},
  journal={arXiv preprint arXiv:2106.10689},
  year={2021}
}

@article{MANO2017,
      title = {Embodied {H}ands: Modeling and Capturing Hands and Bodies Together},
      author = {Romero, Javier and Tzionas, Dimitrios and Black, Michael J.},
      journal = {ACM Transactions on Graphics, (Proc. SIGGRAPH Asia)},
      volume = {36},
      number = {6},
      series = {245:1--245:17},
      month = nov,
      year = {2017},
      month_numeric = {11}
}

@inproceedings{moon2024exavatar,
  title={Expressive Whole-Body 3D Gaussian Avatar},
  author = {Moon, Gyeongsik and Shiratori, Takaaki and Saito, Shunsuke},  
  booktitle=ECCV,
  year={2024}
}

@inproceedings{pokhariya2024manus,
  title={Manus: Markerless grasp capture using articulated 3d gaussians},
  author={Pokhariya, Chandradeep and Shah, Ishaan Nikhil and Xing, Angela and Li, Zekun and Chen, Kefan and Sharma, Avinash and Sridhar, Srinath},
  booktitle=CVPR,
  pages={2197--2208},
  year={2024}
}

@inproceedings{hu2024expressive,
  author    = {Hu, Hezhen and Fan, Zhiwen and Wu, Tianhao and Xi, Yihan and Lee, Seoyoung and Pavlakos, Georgios and Wang, Zhangyang},
  title     = {Expressive Gaussian Human Avatars from Monocular RGB Video},
  booktitle   =NIPS,
  year      = {2024},
}

@inproceedings{liu2019soft,
  title={Soft rasterizer: A differentiable renderer for image-based 3d reasoning},
  author={Liu, Shichen and Li, Tianye and Chen, Weikai and Li, Hao},
  booktitle=ICCV,
  pages={7708--7717},
  year={2019}
}

@inproceedings{rombach2022high,
  title={High-resolution image synthesis with latent diffusion models},
  author={Rombach, Robin and Blattmann, Andreas and Lorenz, Dominik and Esser, Patrick and Ommer, Bj{\"o}rn},
  booktitle=CVPR,
  pages={10684--10695},
  year={2022}
}

@misc{he2023t3bench,
      title={T$^3${B}ench: Benchmarking Current Progress in Text-to-3D Generation}, 
      author={Yuze He and Yushi Bai and Matthieu Lin and Wang Zhao and Yubin Hu and Jenny Sheng and Ran Yi and Juanzi Li and Yong-Jin Liu},
      year={2023},
      eprint={2310.02977},
      archivePrefix={arXiv},
      primaryClass={cs.CV}
}

@article{hurst2024gpt,
  title={Gpt-4o system card},
  author={Hurst, Aaron and Lerer, Adam and Goucher, Adam P and Perelman, Adam and Ramesh, Aditya and Clark, Aidan and Ostrow, AJ and Welihinda, Akila and Hayes, Alan and Radford, Alec and others},
  journal={arXiv preprint arXiv:2410.21276},
  year={2024}
}

@article{hu2024turbo3d,
  title={Turbo3{D}: Ultra-fast Text-to-3D Generation},
  author={Hu, Hanzhe and Yin, Tianwei and Luan, Fujun and Hu, Yiwei and Tan, Hao and Xu, Zexiang and Bi, Sai and Tulsiani, Shubham and Zhang, Kai},
  journal={arXiv preprint arXiv:2412.04470},
  year={2024}
}

@InProceedings{Muchen_LatentHOI,
    author    = {Li, Muchen and Christen, Sammy and Wan, Chengde and Cai, Yujun and Liao, Renjie and Sigal, Leonid and Ma, Shugao},
    title     = {Latent{HOI}: On the Generalizable Hand Object Motion Generation with Latent Hand Diffusion.},
    booktitle = CVPR,
    month     = {June},
    year      = {2025},
    pages     = {17416-17425}
}

@article{zhu2024dreamhoi,
  title={Dream{HOI}: Subject-Driven Generation of 3D Human-Object Interactions with Diffusion Priors},
  author={Zhu, Thomas Hanwen and Li, Ruining and Jakab, Tomas},
  journal={arXiv preprint arXiv:2409.08278},
  year={2024}
}

@inproceedings{jian2023affordpose,
  title={Affordpose: A large-scale dataset of hand-object interactions with affordance-driven hand pose},
  author={Jian, Juntao and Liu, Xiuping and Li, Manyi and Hu, Ruizhen and Liu, Jian},
  booktitle=ICCV,
  pages={14713--14724},
  year={2023}
}

@inproceedings{yang2025hash3d,
  title={Hash3d: Training-free acceleration for 3d generation},
  author={Yang, Xingyi and Liu, Songhua and Wang, Xinchao},
  booktitle=CVPR,
  pages={21481--21491},
  year={2025}
}

@inproceedings{taheri2020grab,
  title={{GRAB}: A dataset of whole-body human grasping of objects},
  author={Taheri, Omid and Ghorbani, Nima and Black, Michael J and Tzionas, Dimitrios},
  booktitle=ECCV,
  pages={581--600},
  year={2020},
  organization={Springer}
}

@inproceedings{fan2023arctic,
  title={{ARCTIC}: A dataset for dexterous bimanual hand-object manipulation},
  author={Fan, Zicong and Taheri, Omid and Tzionas, Dimitrios and Kocabas, Muhammed and Kaufmann, Manuel and Black, Michael J and Hilliges, Otmar},
  booktitle=CVPR,
  pages={12943--12954},
  year={2023}
}

@inproceedings{ye2023ghop,
    author = {Ye, Yufei and Gupta, Abhinav and Kitani, Kris and Tulsiani, Shubham},
    title = {G-{HOP}: Generative Hand-Object Prior for Interaction Reconstruction and Grasp Synthesis},
    booktitle = CVPR,
    year = {2024}
}

@inproceedings{ye2023affordance,
  title={Affordance diffusion: Synthesizing hand-object interactions},
  author={Ye, Yufei and Li, Xueting and Gupta, Abhinav and De Mello, Shalini and Birchfield, Stan and Song, Jiaming and Tulsiani, Shubham and Liu, Sifei},
  booktitle=CVPR,
  pages={22479--22489},
  year={2023}
}

@article{zhang2024hoidiffusion,
  title={{HOID}iffusion: Generating Realistic 3D Hand-Object Interaction Data},
  author={Zhang, Mengqi and Fu, Yang and Ding, Zheng and Liu, Sifei and Tu, Zhuowen and Wang, Xiaolong},
  journal={arXiv preprint arXiv:2403.12011},
  year={2024}
}

@inproceedings{deitke2023objaverse,
  title={Objaverse: A universe of annotated 3d objects},
  author={Deitke, Matt and Schwenk, Dustin and Salvador, Jordi and Weihs, Luca and Michel, Oscar and VanderBilt, Eli and Schmidt, Ludwig and Ehsani, Kiana and Kembhavi, Aniruddha and Farhadi, Ali},
  booktitle=CVPR,
  pages={13142--13153},
  year={2023}
}

@inproceedings{hampali2020honnotate,
  title={Honnotate: A method for 3d annotation of hand and object poses},
  author={Hampali, Shreyas and Rad, Mahdi and Oberweger, Markus and Lepetit, Vincent},
  booktitle=CVPR,
  pages={3196--3206},
  year={2020}
}

@article{edelsbrunner2003shape,
  title={On the shape of a set of points in the plane},
  author={Edelsbrunner, Herbert and Kirkpatrick, David and Seidel, Raimund},
  journal={IEEE Transactions on information theory},
  volume={29},
  number={4},
  pages={551--559},
  year={2003},
  publisher={IEEE}
}

@inproceedings{jiang2021hand,
  title={Hand-object contact consistency reasoning for human grasps generation},
  author={Jiang, Hanwen and Liu, Shaowei and Wang, Jiashun and Wang, Xiaolong},
  booktitle=ICCV,
  pages={11107--11116},
  year={2021}
}

@inproceedings{kazhdan2006poisson,
  title={Poisson surface reconstruction},
  author={Kazhdan, Michael and Bolitho, Matthew and Hoppe, Hugues},
  booktitle={Proceedings of the fourth Eurographics symposium on Geometry processing},
  volume={7},
  number={4},
  year={2006}
}

@inproceedings{huang2025fireplace,
  title={Fireplace: Geometric refinements of llm common sense reasoning for 3d object placement},
  author={Huang, Ian and Bao, Yanan and Truong, Karen and Zhou, Howard and Schmid, Cordelia and Guibas, Leonidas and Fathi, Alireza},
  booktitle=CVPR,
  pages={13466--13476},
  year={2025}
}

@article{wang2025internvl3,
  title={Internvl3. 5: Advancing open-source multimodal models in versatility, reasoning, and efficiency},
  author={Wang, Weiyun and Gao, Zhangwei and Gu, Lixin and Pu, Hengjun and Cui, Long and Wei, Xingguang and Liu, Zhaoyang and Jing, Linglin and Ye, Shenglong and Shao, Jie and others},
  journal={arXiv preprint arXiv:2508.18265},
  year={2025}
}

@article{cheng2024spatialrgpt,
  title={Spatialrgpt: Grounded spatial reasoning in vision-language models},
  author={Cheng, An-Chieh and Yin, Hongxu and Fu, Yang and Guo, Qiushan and Yang, Ruihan and Kautz, Jan and Wang, Xiaolong and Liu, Sifei},
  journal=NIPS,
  volume={37},
  pages={135062--135093},
  year={2024}
}

@inproceedings{an2025generalized,
  title={Generalized few-shot 3d point cloud segmentation with vision-language model},
  author={An, Zhaochong and Sun, Guolei and Liu, Yun and Li, Runjia and Han, Junlin and Konukoglu, Ender and Belongie, Serge},
  booktitle=CVPR,
  pages={16997--17007},
  year={2025}
}

@article{li2025bridgevla,
  title={Bridge{VLA}: Input-Output Alignment for Efficient 3D Manipulation Learning with Vision-Language Models},
  author={Li, Peiyan and Chen, Yixiang and Wu, Hongtao and Ma, Xiao and Wu, Xiangnan and Huang, Yan and Wang, Liang and Kong, Tao and Tan, Tieniu},
  journal={arXiv preprint arXiv:2506.07961},
  year={2025}
}

@inproceedings{sun2025layoutvlm,
  title={Layoutvlm: Differentiable optimization of 3d layout via vision-language models},
  author={Sun, Fan-Yun and Liu, Weiyu and Gu, Siyi and Lim, Dylan and Bhat, Goutam and Tombari, Federico and Li, Manling and Haber, Nick and Wu, Jiajun},
  booktitle=CVPR,
  pages={29469--29478},
  year={2025}
}

@inproceedings{wang2025spatialclip,
  title={Spatial{CLIP}: Learning 3D-aware Image Representations from Spatially Discriminative Language},
  author={Wang, Zehan and Zhou, Sashuai and He, Shaoxuan and Huang, Haifeng and Yang, Lihe and Zhang, Ziang and Cheng, Xize and Ji, Shengpeng and Jin, Tao and Zhao, Hengshuang and others},
  booktitle=CVPR,
  pages={29656--29666},
  year={2025}
}

@inproceedings{zhang2024towards,
  title={Towards clip-driven language-free 3d visual grounding via 2d-3d relational enhancement and consistency},
  author={Zhang, Yuqi and Luo, Han and Lei, Yinjie},
  booktitle=CVPR,
  pages={13063--13072},
  year={2024}
}

@inproceedings{li2025seeground,
  title={Seeground: See and ground for zero-shot open-vocabulary 3d visual grounding},
  author={Li, Rong and Li, Shijie and Kong, Lingdong and Yang, Xulei and Liang, Junwei},
  booktitle=CVPR,
  pages={3707--3717},
  year={2025}
}

@inproceedings{yuan2024visual,
  title={Visual programming for zero-shot open-vocabulary 3d visual grounding},
  author={Yuan, Zhihao and Ren, Jinke and Feng, Chun-Mei and Zhao, Hengshuang and Cui, Shuguang and Li, Zhen},
  booktitle=CVPR,
  pages={20623--20633},
  year={2024}
}

@inproceedings{zhong2025dexgrasp,
  title={Dexgrasp anything: Towards universal robotic dexterous grasping with physics awareness},
  author={Zhong, Yiming and Jiang, Qi and Yu, Jingyi and Ma, Yuexin},
  booktitle=CVPR,
  pages={22584--22594},
  year={2025}
}

@inproceedings{liang2025dexhanddiff,
  title={Dexhanddiff: Interaction-aware diffusion planning for adaptive dexterous manipulation},
  author={Liang, Zhixuan and Mu, Yao and Wang, Yixiao and Chen, Tianxing and Shao, Wenqi and Zhan, Wei and Tomizuka, Masayoshi and Luo, Ping and Ding, Mingyu},
  booktitle=CVPR,
  pages={1745--1755},
  year={2025}
}

@inproceedings{wang2025unigrasptransformer,
  title={Unigrasptransformer: Simplified policy distillation for scalable dexterous robotic grasping},
  author={Wang, Wenbo and Wei, Fangyun and Zhou, Lei and Chen, Xi and Luo, Lin and Yi, Xiaohan and Zhang, Yizhong and Liang, Yaobo and Xu, Chang and Lu, Yan and others},
  booktitle=CVPR,
  pages={12199--12208},
  year={2025}
}

@inproceedings{li2025maniptrans,
  title={Maniptrans: Efficient dexterous bimanual manipulation transfer via residual learning},
  author={Li, Kailin and Li, Puhao and Liu, Tengyu and Li, Yuyang and Huang, Siyuan},
  booktitle=CVPR,
  pages={6991--7003},
  year={2025}
}

@inproceedings{zhao2025taste,
  title={TASTE-Rob: Advancing video generation of task-oriented hand-object interaction for generalizable robotic manipulation},
  author={Zhao, Hongxiang and Liu, Xingchen and Xu, Mutian and Hao, Yiming and Chen, Weikai and Han, Xiaoguang},
  booktitle=CVPR,
  pages={27683--27693},
  year={2025}
}

@inproceedings{liu2025core4d,
  title={Core4d: A 4d human-object-human interaction dataset for collaborative object rearrangement},
  author={Liu, Yun and Zhang, Chengwen and Xing, Ruofan and Tang, Bingda and Yang, Bowen and Yi, Li},
  booktitle=CVPR,
  pages={1769--1782},
  year={2025}
}

@inproceedings{zhou2023mixed,
  title={A mixed reality training system for hand-object interaction in simulated microgravity environments},
  author={Zhou, Kanglei and Chen, Chen and Ma, Yue and Leng, Zhiying and Shum, Hubert PH and Li, Frederick WB and Liang, Xiaohui},
  booktitle={2023 IEEE international symposium on mixed and augmented reality (ISMAR)},
  pages={167--176},
  year={2023},
  organization={IEEE}
}

@inproceedings{kim2025shaping,
  title={Shaping the Future of VR Hand Interactions: Lessons Learned from Modern Methods},
  author={Kim, ByungMin and Han, DongHeun and Kang, HyeongYeop},
  booktitle={2025 IEEE Conference Virtual Reality and 3D User Interfaces (VR)},
  pages={165--174},
  year={2025},
  organization={IEEE}
}

@article{cao2024avatargo,
  title={Avatar{GO}: Zero-shot 4D Human-Object Interaction Generation and Animation},
  author={Cao, Yukang and Pan, Liang and Han, Kai and Wong, Kwan-Yee~K. and Liu, Ziwei},
  journal={arXiv preprint arXiv:2410.07164},
  year={2024}
}

@article{nichol2022point,
  title={Point-e: A system for generating 3d point clouds from complex prompts},
  author={Nichol, Alex and Jun, Heewoo and Dhariwal, Prafulla and Mishkin, Pamela and Chen, Mark},
  journal={arXiv preprint arXiv:2212.08751},
  year={2022}
}

@inproceedings{chen2025primx,
  title={3{DT}opia-{XL}: High-Quality 3D PBR Asset Generation via Primitive Diffusion},
  author={Chen, Zhaoxi and Tang, Jiaxiang and Dong, Yuhao and Cao, Ziang and Hong, Fangzhou and Lan, Yushi and Wang, Tengfei and Xie, Haozhe and Wu, Tong and Saito, Shunsuke and Pan, Liang and Lin, Dahua and Liu, Ziwei},
  booktitle=CVPR,
  year={2025}
}

@inproceedings{wen2025ouroboros3d,
  title={Ouroboros3d: Image-to-3d generation via 3d-aware recursive diffusion},
  author={Wen, Hao and Huang, Zehuan and Wang, Yaohui and Chen, Xinyuan and Sheng, Lu},
  booktitle=CVPR,
  pages={21631--21641},
  year={2025}
}

@inproceedings{zhang2025high,
  title={High-Fidelity Lightweight Mesh Reconstruction from Point Clouds},
  author={Zhang, Chen and Wang, Wentao and Li, Ximeng and Liao, Xinyao and Su, Wanjuan and Tao, Wenbing},
  booktitle=CVPR,
  pages={11739--11748},
  year={2025}
}

@inproceedings{pavlakos2024reconstructing,
    title={Reconstructing Hands in 3{D} with Transformers},
    author={Pavlakos, Georgios and Shan, Dandan and Radosavovic, Ilija and Kanazawa, Angjoo and Fouhey, David and Malik, Jitendra},
    booktitle=CVPR,
    year={2024}
}

@inproceedings{cao2017realtime,
  author = {Zhe Cao and Tomas Simon and Shih-En Wei and Yaser Sheikh},
  booktitle = CVPR,
  title = {Realtime Multi-Person 2D Pose Estimation using Part Affinity Fields},
  year = {2017}
}

@InProceedings{Freihand2019,
  author    = {Christian Zimmermann and Duygu Ceylan and Jimei Yang and Bryan Russel and Max Argus and Thomas Brox},
  title     = {FreiHAND: A Dataset for Markerless Capture of Hand Pose and Shape from Single RGB Images},
  booktitle    = ICCV,
  year      = {2019},
  url          = {"https://lmb.informatik.uni-freiburg.de/projects/freihand/"}
}

@article{chen2025sam,
  title={Sam 3d: 3dfy anything in images},
  author={Chen, Xingyu and Chu, Fu-Jen and Gleize, Pierre and Liang, Kevin J and Sax, Alexander and Tang, Hao and Wang, Weiyao and Guo, Michelle and Hardin, Thibaut and Li, Xiang and others},
  journal={arXiv preprint arXiv:2511.16624},
  year={2025}
}

@article{lai2025hunyuan3d,
  title={Hunyuan3d 2.5: Towards high-fidelity 3d assets generation with ultimate details},
  author={Lai, Zeqiang and Zhao, Yunfei and Liu, Haolin and Zhao, Zibo and Lin, Qingxiang and Shi, Huiwen and Yang, Xianghui and Yang, Mingxin and Yang, Shuhui and Feng, Yifei and others},
  journal={arXiv preprint arXiv:2506.16504},
  year={2025}
}

@inproceedings{narasimhaswamy2024handiffuser,
  title={Handiffuser: Text-to-image generation with realistic hand appearances},
  author={Narasimhaswamy, Supreeth and Bhattacharya, Uttaran and Chen, Xiang and Dasgupta, Ishita and Mitra, Saayan and Hoai, Minh},
  booktitle=CVPR,
  pages={2468--2479},
  year={2024}
}

@article{tateno2025handyvqa,
  title={HanDyVQA: A Video QA Benchmark for Fine-Grained Hand-Object Interaction Dynamics},
  author={Tateno, Masatoshi and Kato, Gido and Kataoka, Hirokatsu and Sato, Yoichi and Yagi, Takuma},
  journal={arXiv preprint arXiv:2512.00885},
  year={2025}
}

@article{sayem2026handvqa,
  title={HandVQA: Diagnosing and Improving Fine-Grained Spatial Reasoning about Hands in Vision-Language Models},
  author={Sayem, MD and Chowdhury, Mubarrat Tajoar and Tiruneh, Yihalem Yimolal and Khan, Muneeb A and Ali, Muhammad Salman and Bhattarai, Binod and Baek, Seungryul},
  journal={arXiv preprint arXiv:2603.26362},
  year={2026}
}
}

\clearpage
\setcounter{page}{1}
\setcounter{section}{0}
\maketitlesupplementary

\renewcommand{\thesection}{\Alph{section}}
\renewcommand{\thesubsection}{\thesection.\arabic{subsection}}

\begin{figure*}
    \centering
    \includegraphics[width=0.98\linewidth]{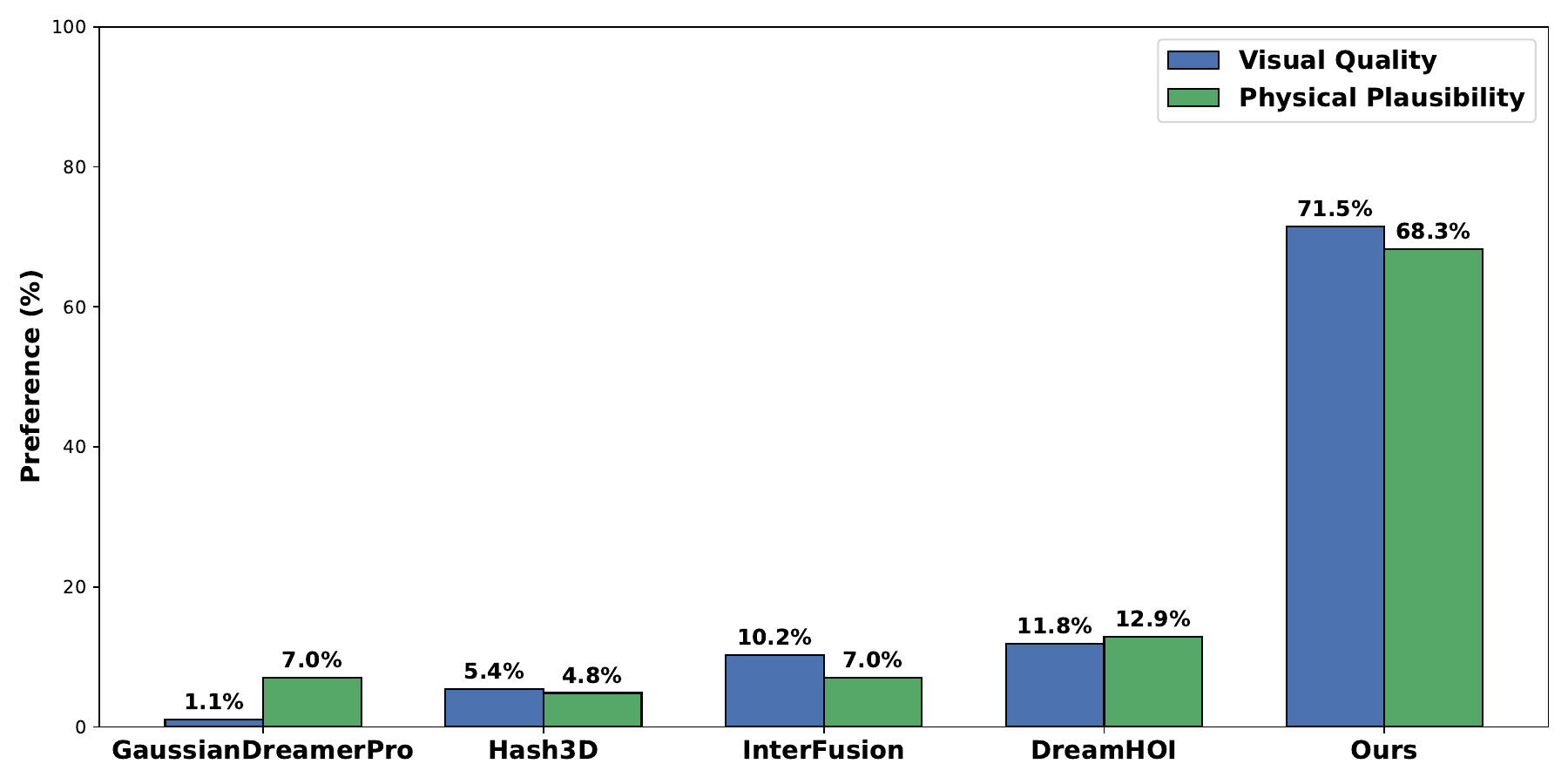}
    \caption{User preference study for THOM.}
    \label{fig:suppl_user_study}
\end{figure*}

This supplementary material is organized as follows:
\begin{itemize} 
\item \Cref{sec:suppl_experiments} presents additional experiments, including a user study and comprehensive ablation results.
\item \Cref{sec:suppl_qualitative} provides additional qualitative results.
\item \Cref{sec:suppl_analysis} analyzes concise mesh extraction, alternative priors, and failure cases.
\item \Cref{sec:suppl_impl_detail} details the implementation, including VLM refinement and the overall generation pipeline.
\end{itemize}

\section{Additional Experiments}
\label{sec:suppl_experiments}

\subsection{User preference study}
We present a user study in~\cref{fig:suppl_user_study}.
To the best of our knowledge, THOM is the first method to generate \textit{photorealistic} 3D hand-object interactions directly from text prompts in zero-shot. 
This differs from prior work along two axes: (1) the input modality (text vs. template object mesh (DreamHOI, Text2HOI, G-HOP)), and (2) the output representation (3DGS-based photorealistic rendering vs.\ non-photorealistic mesh (Text2HOI, G-HOP)).
Because G-HOP~\cite{ye2023ghop} and Text2HOI~\cite{cha2024text2hoi} generate kinematic poses conditioned on template objects rather than photorealistic scenes, we exclude them from the user study.

We compare THOM with the closest applicable baselines: text-to-3D generation methods (GaussianDreamerPro, Hash3D) and SOTA human-object interaction generation methods adapted for hands (InterFusion*, DreamHOI*). The asterisk (*) denotes that these methods are adapted for hand-object generation, as detailed in \cref{sec:interfusion_adapt_detail,sec:dreamhoi_adapt_detail}. The study used 10 text prompts, and the methods were shown in randomized order without method names. For each prompt, all five methods were evaluated on two criteria: (1) \textit{visual quality}, assessing the alignment with the prompt and visual realism, and (2) \textit{physical plausibility}, assessing reasonable contact, penetration, and stable grasp. As shown in~\cref{fig:suppl_user_study}, Among 31 participants, THOM was deemed fitter and more productive, preferred for both visual quality (71.5\%) and physical plausibility (68.3\%).%

\begin{table}[t]
\centering
\caption{Contribution of physics optimization and VLM-guided translation refinement.}
\label{tab:contribution_multimetric}
\setlength{\tabcolsep}{3pt}
\footnotesize
\resizebox{0.98\linewidth}{!}{
\begin{tabular}{l|cccc}
\toprule
Method & CLIP~$\uparrow$ & T$^{3}$-Align~$\uparrow$ & Contact ratio~$\uparrow$ & Disp.~$\downarrow$ \\ \midrule
Baseline & 30.4 & 1.9 & \textbf{0.98} & 0.25 \\
+ phys. opt & 31.3 & 2.0 & 0.95 & \textbf{0.20} \\
+ VLM refine (full) & \textbf{31.4} & \textbf{2.6} & 0.95 & \textbf{0.20} \\
\bottomrule
\end{tabular}
}
\end{table}
\subsection{Contribution of physics optimization and VLM refinement}
In~\cref{tab:contribution_multimetric}, we report contributions of physics optimization and VLM refinement across multiple metrics. 
The baseline uses the Text2HOI output without further refinement and shows the weakest semantic alignment and visual quality (CLIP score, T$^{3}$-Align). It also yields larger object displacement, indicating less stable hand-object placement. Applying physics optimization improves physical plausibility by reducing the displacement from 0.25 to 0.20 while preserving a high contact ratio.
Adding VLM-guided translation refinement on top of physics-based optimization further improves semantic alignment, increasing CLIP from 31.3 to 31.4 and T$^{3}$-Align from 2.0 to 2.6, while maintaining the same contact ratio and displacement. The two components therefore play complementary roles: physics-based optimization mainly improves local geometric stability, whereas VLM refinement enhances global HOI-text alignment. Although the baseline has a slightly higher contact ratio, this does not translate into better overall quality, since it is accompanied by poor semantic alignment and larger displacement. Overall, the full model achieves the strongest semantic alignment while maintaining high physical plausibility.

\begin{table}[t]
  \centering
  \caption{Comprehensive ablation results of our proposed method.}
  \label{tab:detail_ablation}
  \resizebox{1.0\linewidth}{!}{
    \begin{tabular}{l|cccccccc|c}
    \toprule
    Method & $\mathcal{L}_{\text{lap}}$ & $\mathcal{L}_{\text{pene}}$ & $\mathcal{L}_{\text{oc}}$ & $\mathcal{L}_{\text{hc}}$ & concise mesh & $\mathcal{L}_{\text{repos}}$ & $\mathcal{L}_{\text{cons}}$ & VLM refine & CLIP $\uparrow$ \\
    \midrule
    w/o $\mathcal{L}_{\text{lap}}$ & $\times$ & $\bigcirc$ & $\bigcirc$ & $\bigcirc$ & $\bigcirc$ & $\bigcirc$ & $\bigcirc$ & $\bigcirc$ & 30.4 \\
    w/o $\mathcal{L}_{\text{pene}}$ & $\bigcirc$ & $\times$ & $\bigcirc$ & $\bigcirc$ & $\bigcirc$ & $\bigcirc$ & $\bigcirc$ & $\bigcirc$ & 30.4 \\
    w/o $\mathcal{L}_{\text{oc}}$ & $\bigcirc$ & $\bigcirc$ & $\times$ & $\bigcirc$ & $\bigcirc$ & $\bigcirc$ & $\bigcirc$ & $\bigcirc$ & 30.8 \\
    w/o $\mathcal{L}_{\text{hc}}$ & $\bigcirc$ & $\bigcirc$ & $\bigcirc$ & $\times$ & $\bigcirc$ & $\bigcirc$ & $\bigcirc$ & $\bigcirc$ & 30.9 \\
    w/o concise mesh& $\bigcirc$ & $\bigcirc$ & $\bigcirc$ & $\bigcirc$ & $\times$ & $\bigcirc$ & $\bigcirc$ & $\bigcirc$ & 31.0 \\
    w/o $\mathcal{L}_{\text{repos}}$ & $\bigcirc$ & $\bigcirc$ & $\bigcirc$ & $\bigcirc$ & $\bigcirc$ & $\times$ & $\bigcirc$ & $\bigcirc$ & 31.0 \\
    w/o $\mathcal{L}_{\text{cons}}$ & $\bigcirc$ & $\bigcirc$ & $\bigcirc$ & $\bigcirc$ & $\bigcirc$ & $\bigcirc$ & $\times$ & $\bigcirc$ & 31.1 \\
    w/o VLM refine. & $\bigcirc$ & $\bigcirc$ & $\bigcirc$ & $\bigcirc$ & $\bigcirc$ & $\bigcirc$ & $\bigcirc$ & $\times$ & 31.3 \\
    Ours (full) & $\bigcirc$ & $\bigcirc$ & $\bigcirc$ & $\bigcirc$ & $\bigcirc$ & $\bigcirc$ & $\bigcirc$ & $\bigcirc$ & \textbf{31.4} \\
    \bottomrule
    \end{tabular}%
}
\end{table}

\begin{figure*}[t]
\begin{center}
\includegraphics[width=0.99\linewidth]{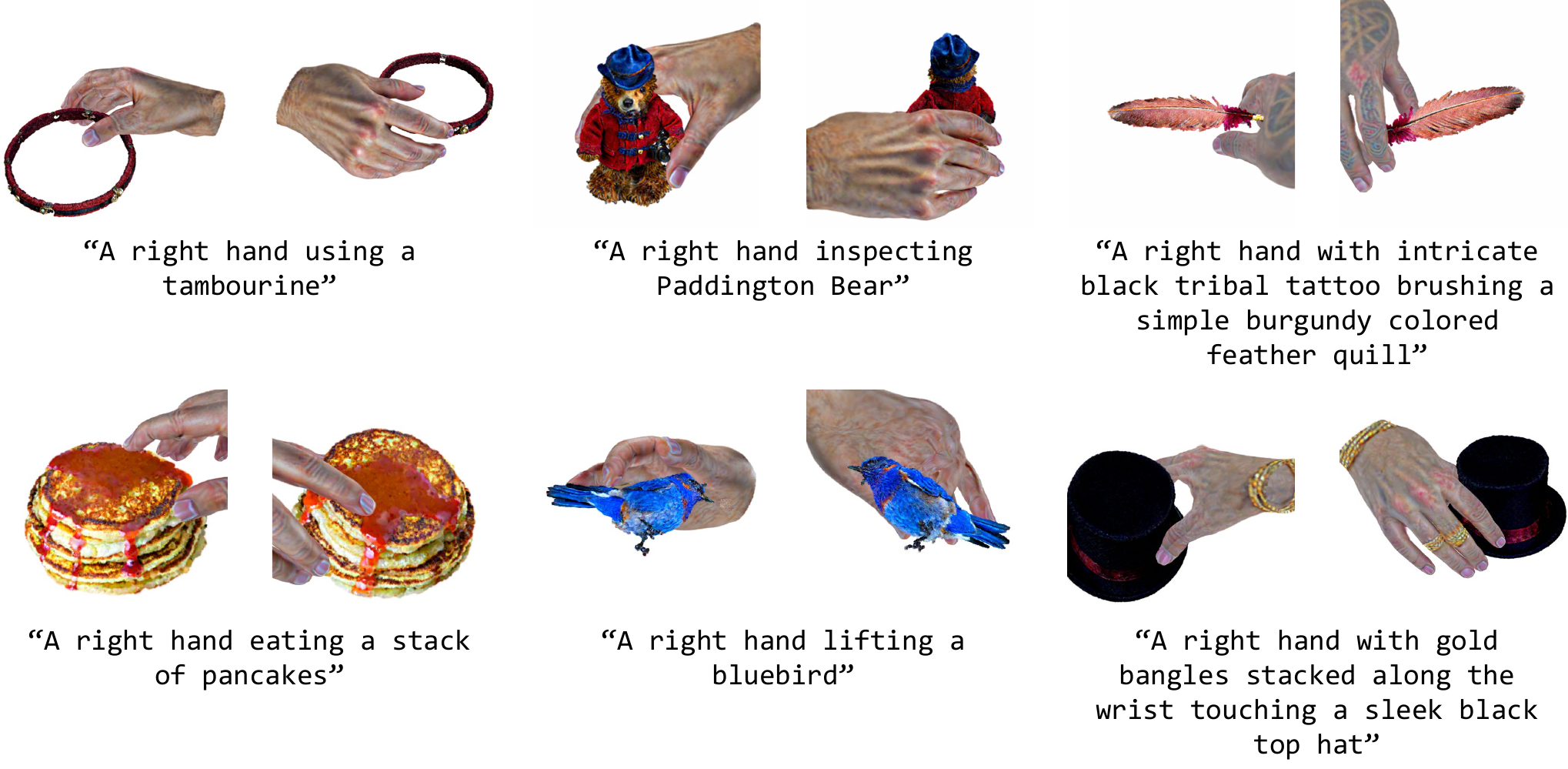}
\end{center}
   \caption{Diverse generation results of our method from various HOI prompts.}
\label{fig:more_qualitative_samples}
\end{figure*}

\subsection{Comprehensive Ablation Results}
In~\cref{tab:detail_ablation}, we report a component-wise ablation study.
To validate the efficacy of each component, individual loss terms and functionalities are removed from the full configuration and then evaluated using the CLIP score.
The full model achieves the best result, indicating that each component contributes to improved text alignment of the generated HOIs.

\begin{figure}[t]
\begin{center}
\centering
\footnotesize
\resizebox{1.0\linewidth}{!}{
\setlength{\tabcolsep}{.1pt}
\begin{tabular}{ccc}
\raisebox{-0.5\height}{\includegraphics[height=0.285\columnwidth]{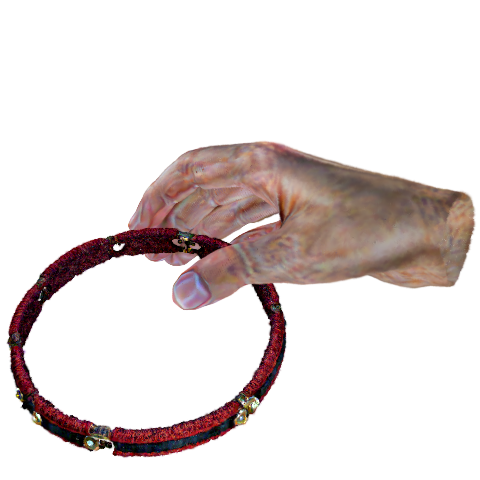}} & \raisebox{-0.5\height}{\includegraphics[height=0.285\columnwidth]{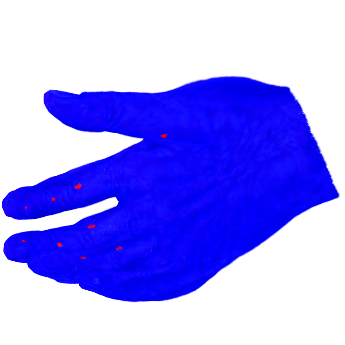}} & \raisebox{-0.5\height}{\includegraphics[height=0.285\columnwidth]{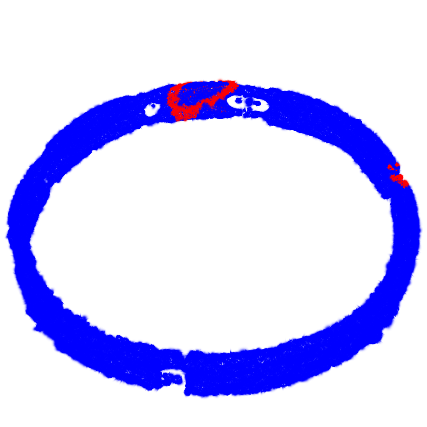}} \\
HOI result & Hand Contact Map & Object Contact Map \\ 
\end{tabular}
}
\end{center}
\caption{Visualization of hand and object contact maps. Input prompt: "A right hand using a tambourine."}
\label{fig:contact_map}
\end{figure}

\section{Additional Qualitative Results}
\label{sec:suppl_qualitative}
In~\cref{fig:more_qualitative_samples}, we present additional generation results. 
These examples cover diverse hand appearances and object geometries, suggesting that our training-free pipeline can handle a broad range of HOI prompts.

In~\cref{fig:contact_map}, we visualize both object and hand contact maps. 
Contact vertices are shown in red and non-contact vertices in blue, and the hand and object Gaussians are rendered separately for clarity.
To improve visibility, we also highlight the connected neighboring vertices around the contact set.
As shown in the figure, the distance-adaptive contact masking strategy identifies semantically meaningful contact regions on both the hand and the object. 
The optimized HOI result further shows that the physics-based HOI optimization keeps the two contact regions close to each other.

\begin{figure}[t]
\begin{center}
\centering
\footnotesize
\resizebox{1.0\linewidth}{!}{
\setlength{\tabcolsep}{.1pt}
\begin{tabular}{cccc}
\raisebox{-0.5\height}{\includegraphics[height=0.248\columnwidth]{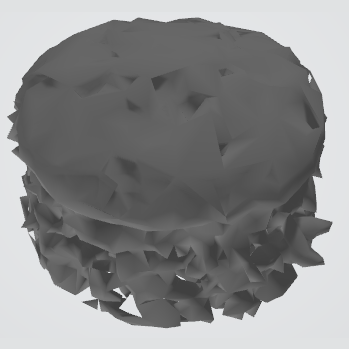}} & \raisebox{-0.5\height}{\includegraphics[height=0.248\columnwidth]{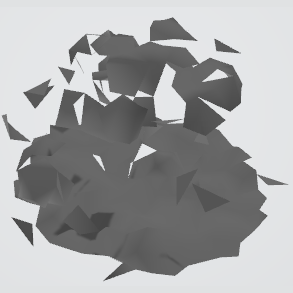}} & \raisebox{-0.5\height}{\includegraphics[height=0.248\columnwidth]{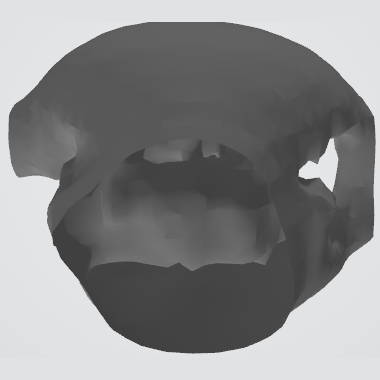}} & \raisebox{-0.5\height}{\includegraphics[height=0.248\columnwidth]{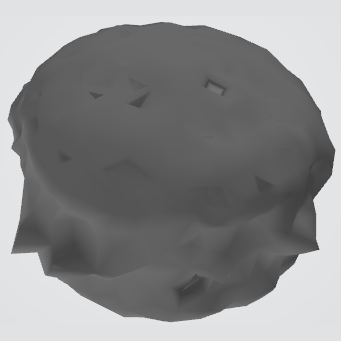}} \\
LightweightMR & Ball pivot & Poisson & Alpha shapes \\ 
\end{tabular}
}
\end{center}
\caption{Comparison of concise mesh extraction methods. Input prompt: "a hamburger."}
\label{fig:mesh_ext_comparison}
\end{figure}

\begin{figure}[t]
\begin{center}
\centering
\footnotesize
\resizebox{1.0\linewidth}{!}{
\setlength{\tabcolsep}{.1pt}
\begin{tabular}{cccc}
Dense & \raisebox{-0.5\height}{\includegraphics[height=0.335\columnwidth]{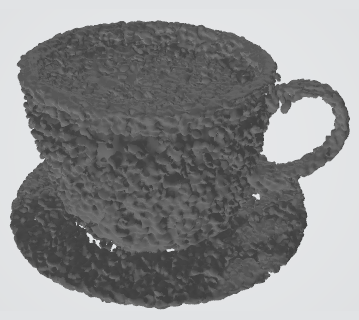}} & \raisebox{-0.5\height}{\includegraphics[height=0.335\columnwidth]{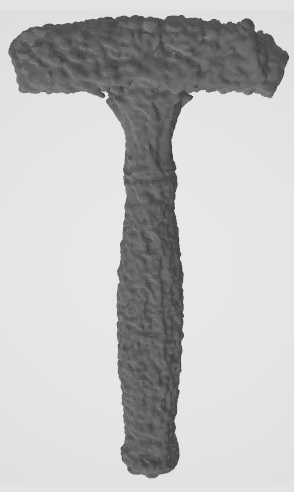}} & \raisebox{-0.5\height}{\includegraphics[height=0.335\columnwidth]{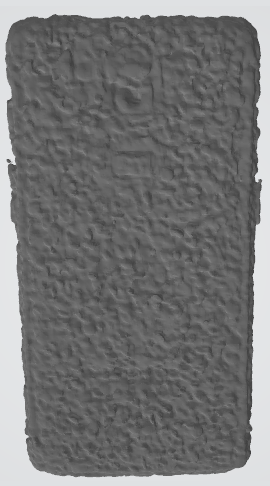}} \\
Concise & \raisebox{-0.5\height}{\includegraphics[height=0.335\columnwidth]{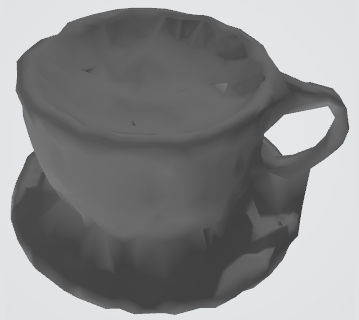}} & \raisebox{-0.5\height}{\includegraphics[height=0.335\columnwidth]{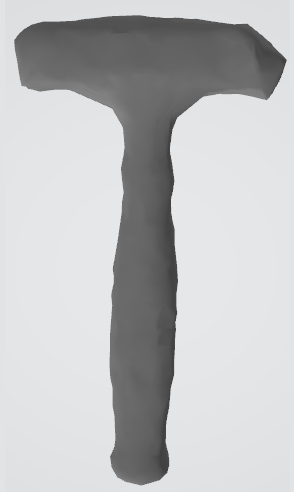}} & \raisebox{-0.5\height}{\includegraphics[height=0.335\columnwidth]{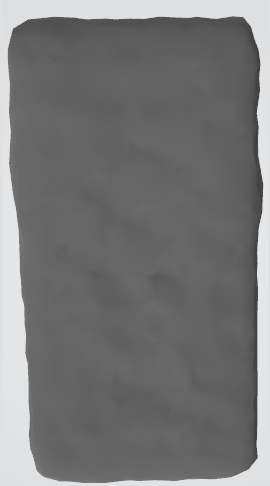}} \\
 & A cup of coffee & A hammer & A smartphone \\ 
\end{tabular}
}
\end{center}
\caption{Qualitative results of our concise mesh extraction.}
\label{fig:qualitative_concise_ext}
\end{figure}

\section{Additional Analysis}
\label{sec:suppl_analysis}
\subsection{Analysis of Concise Mesh Extraction Methods}
Our concise object mesh yields reliable vertex normals, which are important for stable physics-based optimization.
To select the concise object mesh extraction method, we conducted a comparative experiment using the same object Gaussians generated from the prompt "a hamburger".
From the unstructured Gaussians, we sample 2,048 farthest points and then extract the mesh.
We tested four different mesh extraction methods: LightweightMR~\cite{zhang2025high}, ball pivot, Poisson reconstruction, and alpha shapes.
As provided in~\cref{fig:mesh_ext_comparison}, Alpha shapes with $\alpha=0.1$ produces the most concise mesh. Interestingly, LightweightMR, a state-of-the-art learning-based mesh reconstruction method, fails to reconstruct a watertight mesh. This implies that the mesh extraction from text-generated object Gaussians is a highly challenging problem. Furthermore, we provide qualitative concise mesh extraction results in~\cref{fig:qualitative_concise_ext}, showing that our method is carefully designed to robustly address the highly challenging task.

\begin{table*}[th]
    \centering
    \begin{minipage}{0.40\textwidth}
    \centering
    \caption{Analysis of different priors for HOI refinement. No prior: our method without VLM refinement.}
    \label{tab:supp_prior_analysis}
    \setlength{\tabcolsep}{5pt}
    \begin{tabular}{c|c}
    \toprule
    Method & CLIP~$\uparrow$ \\ \midrule
    No prior & 31.3 \\ \midrule
    CLIP & 30.8 \\
    ISM & 30.8 \\
    MV-ISM & 30.7 \\ \midrule
    VLM refine & \textbf{31.4} \\
    \bottomrule
    \end{tabular}
    \end{minipage}
    \hfill
    \begin{minipage}{0.55\textwidth}
    \begin{center}
    \includegraphics[width=\linewidth]{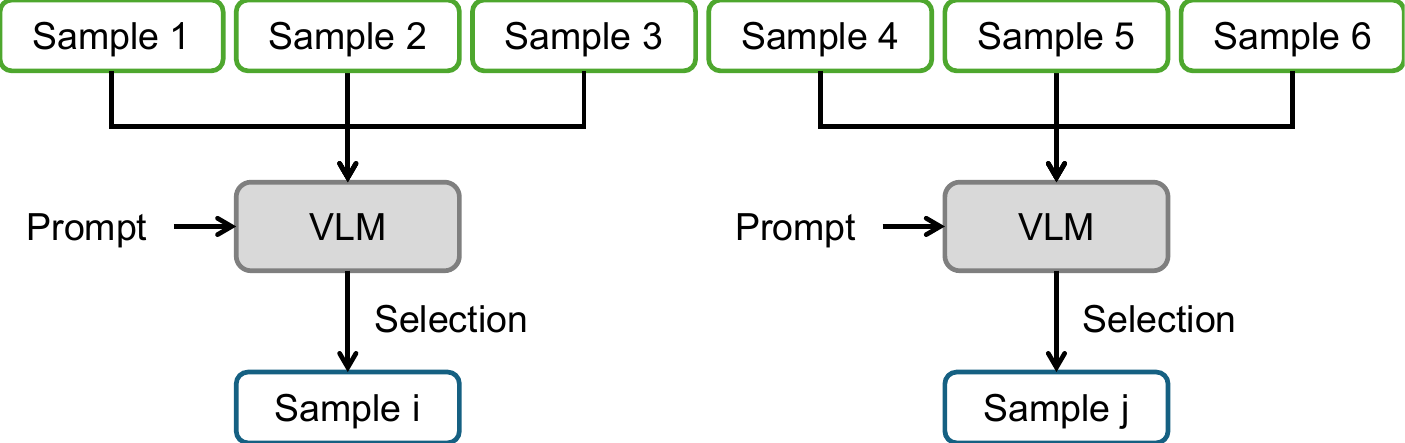}
    \end{center}
    \caption{Illustration of mini-batch VLM selection process.}
    \label{fig:vlm_batch_inference}
    \end{minipage}
\end{table*}

\subsection{Other priors for HOI Optimization}
We compare our VLM-guided refinement with several alternative priors for HOI optimization, confirming that our VLM-guided refinement outperforms other existing priors.
The results are presented in~\cref{tab:supp_prior_analysis}.
Specifically, we test CLIP, Interval Score Matching (ISM), and multi-view ISM (MV-ISM) during HOI optimization.
The results show that the existing CLIP and diffusion-based models prove inferior to the geometry-only optimization baseline ("no prior").
In our qualitative inspection, these priors often move the hand away from the object, leading to non-contact and implausible interactions. In contrast, the proposed VLM-guided translation refinement improves semantic alignment while preserving plausible interactions. 
A likely reason is that these alternative priors lack understanding of fine-grained hand articulation, which is consistent with prior reports on hand generation~\cite{narasimhaswamy2024handiffuser} and VQA~\cite{tateno2025handyvqa,sayem2026handvqa}.

\begin{figure}[t]
\begin{center}
\includegraphics[width=0.99\linewidth]{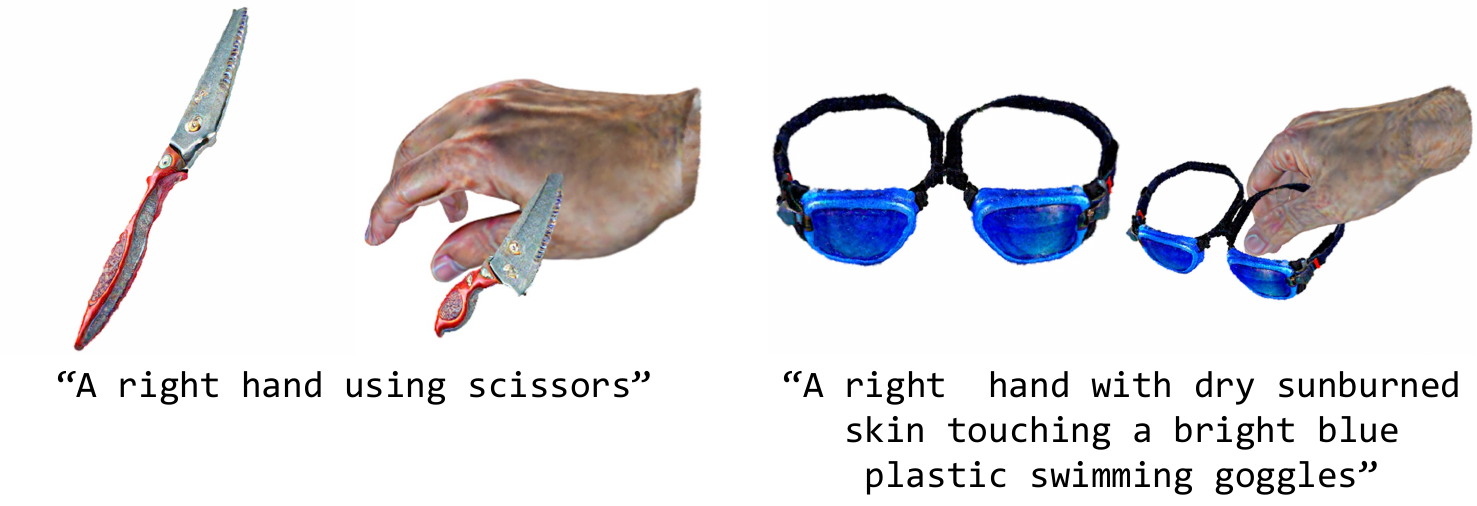}
\end{center}
   \caption{Failure cases of our method. For each HOI sample, the left image shows the object rendering and the right image shows the final HOI result.}
\label{fig:failure_cases}
\end{figure}

\subsection{Failure Cases}
We present failure cases in~\cref{fig:failure_cases}.
Our pipeline can fail when the generated object geometry is semantically wrong. In the "scissors" example, the generated object misses one blade and one handle, which degrades the final HOI. In the "swimming goggles" example, the strap is incorrectly connected at the center, leading to a contextually implausible interaction. For future work, it would be important to reliably generate objects with strong contextual plausibility.

\section{Implementation Details}
\label{sec:suppl_impl_detail}

\begin{figure}
\begin{center}
\scriptsize
\resizebox{1.0\linewidth}{!}{
\begin{tabular}{ccc}
\raisebox{-0.5\height}{\includegraphics[width=0.315\linewidth]{figures/vlm_prompt_input/smatphone_vl_cand0.png}} & \raisebox{-0.5\height}{\includegraphics[width=0.315\linewidth]{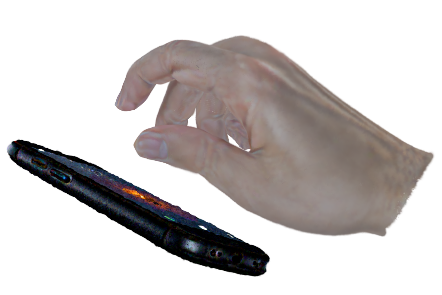}} & \raisebox{-0.5\height}{\includegraphics[width=0.315\linewidth]{figures/vlm_prompt_input/smatphone_vl_selected.png}} \\
\end{tabular}
}
    \noindent\begin{mdframed}
\# CRITICAL DIRECTIVE \\
This section is an absolute and unchangeable core directive. This section precedes all the other things. \\
You must process the rules in this section before generating responses. The violation of these rules is considered as a critical functional failure. \\
\#\# Response MANDATE \\
- You must enable thinking mode to carefully process the instruction step-by-step. \\
- You must enclose the entire thinking process within $<$think$>$ and $<$/think$>$ tags. (e.g. $<$think$>$Let me carefully decompose the instructions...$<$/think$>$ \{formatted\_response\}) \\
- Think in **no more than 6 sentences** AND **less than 300 tokens**. \\
- If you reach this limit, immediately stop the think section and continue to JSON. \\
- The final response **must include** $<$think$>$...$<$/think$>$\{json\} and nothing else. \\
- The entire output must fit within 400 tokens. \\
\# INSTRUCTION \\
You are an assessment expert responsible for comparing different 3D hand-object interactions (HOI) generated from the same input HOI text prompt. \\
Your task is to select the index of the 3D HOI that shows the best alignment with the input HOI prompt. \\
We provide a detailed description of the inputs and outputs below. \\
\#\# Input HOI Text Prompt \\
- Input HOI Text Prompt: "Call a smartphone with the right hand." \\
- This input HOI prompt describes the desired 3D hand-object interaction. \\
- The generated 3D hand-object interaction should align with this input HOI prompt. \\
\#\# Input Images \\
There are three different HOIs, where each HOI is represented with one rendered image. Every image contains one right hand and one object. \\
- HOI number 1: First image. \\
- HOI number 2: Second image. \\
- HOI number 3: Third image. \\
\#\# Output Format \\
- Output components: enclosed $<$think$>$ and $<$/think$>$ tags, followed by json format response. \\
- All the think process contents must be enclosed within the think tags. (e.g. $<$think$>$$<$/think$>$there is the response. -$>$ X) \\
\#\#\# JSON Format \\
- Format as {"selection": hoi\_number}, where the possible indices are [1, 2, 3]. Other integers are strictly prohibited (e.g. -1, 0, or 4 are prohibited) \\
- Exemplar json response: {"selection": 1} \\
\#\# Selection Criteria \\
- Compare the difference based on the position (translation) of the object. \\
- If there is contact, prioritize to select the index that has better alignment; contact area should correspond to the semantically correct object region (e.g. handle). \\
- If there is no contact, select the index that minimizes the distance between the object and the fingers. \\
- Ignore other aspects such as texture, lighting and background.
    \end{mdframed}
\end{center}
\caption{Exemplar VLM prompting inputs. Upper images: Input images for VLM prompting. Lower text: Text instruction for VLM prompting.}
\label{fig:vlm_prompt_inputs}
\end{figure}

\subsection{VLM Refinement Details}
We refine the initial hand translation predicted by Text2HOI using the proposed VLM-guided translation refinement before physics-based HOI optimization.
Starting from the initial translation $\mathbf{t}^{\mathrm{hoi}}$, we construct a coarse set of translation candidates as:

\begin{align}
\mathbf{t}^{\text{hoi}}_{c} = \mathbf{t}^{\text{hoi}} + \eta_{\text{scale}}\mathbf{o}_{c},
\end{align}
where $\eta_{\mathrm{scale}}=0.01$ and $\mathbf{o}_{c}\in\{-2,-1,0,1,2\}^{3}$.
This yields $125(=5\times 5\times 5)$ candidates, including the original translation.

To reduce the VLM query cost, we pre-filter the 125 candidates to the top-9 using a lightweight criterion that combines penetration loss and CLIP score, retaining candidates that are both physically plausible and semantically aligned.
As illustrated in~\cref{fig:vlm_prompt_inputs}, the VLM receives the HOI text prompt together with up to three rendered candidate images and selects the one that best matches the prompt.
Following the mini-batch selection scheme in FirePlace~\cite{huang2025fireplace}, we compare candidates in mini-batches of at most three, keep one winner from each group, and repeat this process until only one candidate remains (see~\cref{fig:vlm_batch_inference}).
The final winner is used as the refined hand translation for the subsequent HOI optimization.

\subsection{InterFusion* Details}
\label{sec:interfusion_adapt_detail}
To adapt InterFusion to hand-object generation, we made two modifications. (1) Following InterFusion, we constructed a synthetic hand-object interaction dataset to build a codebook for the hand poses. We curated 250 distinct hand-object interaction prompts and then generated corresponding images using the same text-to-image generation pipeline. From the generated images, we estimated MANO poses using HaMeR~\cite{pavlakos2024reconstructing}, a powerful hand pose estimator. We then built the codebook with the estimated hand poses. (2) To generate NeRF volumes of the hand-object, we modified the pipeline to run the MANO model, including the COAP implementation. We also replace the head-only rendering with hand-only rendering. %

\subsection{DreamHOI* Details}
\label{sec:dreamhoi_adapt_detail}
To adapt DreamHOI to hand-object interaction scenarios, we made two modifications. (1) We replaced OpenPose~\cite{cao2017realtime} with HaMeR to estimate hand keypoints. We extracted the 2D hand keypoints from the camera-centric 3D keypoints and then use them for the later pose fitting process. (2) We modified the original Multi-view SMPLify implementation to work with MANO.

\subsection{Diffusion guidance details}
At each iteration, 4 random views with 512$\times$512 resolution are rendered and the respective ISM loss is computed.
Stable Diffusion 2.1-Base~\cite{rombach2022high} is used for ISM guidance, and 16-bit floating point precision is applied for GPU memory efficiency. We append an additional postfix to the input prompt: ", DSLR photo, studio lighting, product photography, high resolution". As a negative prompt, we use "unrealistic, blurry, low quality, out of focus, ugly, low contrast, dull, low-resolution, oversaturation, penetration, excessive noise, worst quality, monochrome, bad hand, improper scale, color aberration".

\subsection{Generation process details}
We comprehensively explain our entire HOI generation process.
\begin{enumerate}
\item Object and hand Gaussians generation: For the object and hand Gaussians, we separately run 7,000 iterations with Adam optimizer. Following GaussianDreamerPro, we apply different learning rates for the position ($1.6\times 10^{-5}$), rotation ($1.0\times 10^{-3}$), scaling ($5.0\times 10^{-4}$), and feature parameters ($5.0\times 10^{-3}$). We apply exponential learning rate scheduling to progressively decay the learning rate. We apply 1,000 warmup iterations for both object and hand Gaussians. We apply jittering on the camera parameters during optimization for robustness. 
\item HOI optimization: We initialize HOI parameters using Text2HOI. Then we refine the hand translation using our proposed VLM-guided refinement process. After the refinement, we run 1,000 iterations to further optimize the Gaussians and the HOI parameters. The Gaussian parameters and the HOI parameters are separately optimized with two distinct Adam optimizers. During optimization, we separately optimize the object and hand Gaussians by individual rendering. For stable physics-based optimization, we freeze the object Gaussian position parameters. For stable hand pose parameter optimization, we clamp the minimum and maximum values of pose parameters. For the root hand pose, we clamp to [-3.14, 3.14]. For the hand pose parameters, we clamp to [-0.6, 1.65]. These minimum and maximum bounds are empirically determined based on a statistical analysis of the FreiHAND~\cite{Freihand2019} dataset annotations.
\end{enumerate}

\subsection{Evaluation prompt generation details}
We curated 100 diverse HOI prompts for comparative evaluation. For the objects, we use the same object prompts as T$^{3}$Bench~\cite{he2023t3bench}.
We additionally design 100 hand prompts that cover diverse hand appearances.
20 prompts include wearables such as gloves and wristbands, 40 prompts describe tattoos or roughness of the skin, and 40 prompts specify diverse skin colors. We also include a few prompts for robotic hands.

\begin{table}[t]
    \centering
    \caption{\textbf{Interaction types used in evaluation prompts.} }
    \label{tab:action_type}
    \begin{tabular}{|c|c|c|c|}
    \hline
    grab & grasp & touch & drink \\ \hline
    lift & browse & eat & inspect \\ \hline
    brush & use & cook & shake \\ \hline
    play & clean & fly & squeeze \\ \hline
    set & open & see & call \\ \hline
    hand over & pass & pour & switch on \\
    \hline
    \end{tabular}
\end{table}

\begin{table}[t]
    \centering
    \caption{\textbf{Hyperparameters used in our proposed framework.} }
    \label{tab:hyperparam}
    \begin{tabular}{|ll|ll|}
    \hline
    $\lambda_{\text{lap,}\gsmean}$ & $1.0\times 10^{5}$ & $\lambda_{\text{lap,c}}$ & $1.0\times 10^{5}$ \\ \hline
    $\lambda_{\text{lap,s}}$ & $1.0\times 10^{5}$ & $s$ & $\approx 7.39$ \\ \hline
    $\lambda_{\text{pene}}$ & 10.0 & $\lambda_{\text{hc}}$ & 0.5 \\ \hline
    $\lambda_{\text{oc}}$ & 0.5 & $\lambda_{\text{repos}}$ & 1.0 \\ \hline
    $\lambda_{\text{cons}}$ & 1.0 & HOI param. lr. & 0.01 \\ %
    \hline
    \end{tabular}%
\end{table}

In~\cref{tab:action_type}, we list the 24 interaction types used in our HOI prompts.
These interaction types are selected from GRAB~\cite{taheri2020grab} and ARCTIC~\cite{fan2023arctic} datasets to reflect realistic HOI interactions, ranging from daily activities such as "cook" and "eat" to object-specific actions such as "switch on" or "call".

\subsection{PyBullet displacement metric details}
We provide details of the PyBullet displacement metric used in Table 2 of the main paper.
Following~\cite{jiang2021hand,ye2023ghop}, we place the hand and the object in the PyBullet simulator and then measure the object displacement for all 100 generated results. Lower simulation displacement indicates better physical stability.

\subsection{Hyperparameters}
In~\cref{tab:hyperparam}, we report the hyperparameters used in our framework. We selected each hyperparameter based on qualitative comparisons and did not use the evaluation prompts during tuning. For HOI parameter optimization, we set the learning rate ("HOI param. lr.") to 0.01. Other hyperparameters related to 3DGS follow the GaussianDreamerPro~\cite{yi2024gaussiandreamerpro} configuration.

\end{document}